\documentclass{article}
\usepackage{natbib}
\usepackage[utf8]{inputenc}
\usepackage{amsmath, bm}
\usepackage{amssymb}
\usepackage{mathtools}
\usepackage{amsthm}
\usepackage{thmtools}
\usepackage{thm-restate}
\usepackage{caption}
\usepackage{subcaption}
\usepackage{url}
\usepackage{listings}
\usepackage{fullpage}
\usepackage[dvipsnames]{xcolor}
\usepackage{xspace}
\usepackage{relsize}
\usepackage{hyperref}
\usepackage{float}
\newfloat{algorithm}{t}{lop}
\usepackage{algorithmicx}
\usepackage[indLines=false]{algpseudocodex}
\usepackage[bibliography=common]{apxproof}
\usepackage{enumitem}

\setlist[description]{font=\normalfont\itshape\textbullet\space}

\usepackage{float}
\newfloat{algorithm}{t}{lop}

\floatname{algorithm}{Algorithm}

\hypersetup{
	colorlinks=true,
	linkcolor=blue,
	filecolor=magenta,      
	urlcolor=blue,
	citecolor=blue,
	pdftitle={Overleaf Example},
	pdfpagemode=FullScreen,
}

\title{On Provable Copyright Protection for Generative Models}
\author{Nikhil Vyas\thanks{Harvard School of Engineering and Applied Sciences, \texttt{nikhil@g.harvard.edu}.} 
	\and Sham Kakade\thanks{Harvard School of Engineering and Applied Sciences and Kempner Institute for the Study of Natural and Artificial Intelligence, \texttt{sham@seas.harvard.edu}.}   
	\and Boaz Barak\thanks{Harvard School of Engineering and Applied Sciences, \texttt{b@boazbarak.org}.}}

\newtheoremrep{theorem}{Theorem}[section]
\newtheoremrep{lemma}[theorem]{Lemma}
\newtheoremrep{corollary}[theorem]{Corollary}

\theoremstyle{definition}
\newtheorem{definition}[theorem]{Definition}
\newtheorem{example}[theorem]{Example}

\usepackage{mdframed}

\def \eps {{\varepsilon}}

\newcommand{\E}{\mathbb{E}}

\def\draft{1}

\ifnum\draft=1
\newcommand{\authnote}[3]{\textsf{\color{#3}  $\ll$ #1: { #2} $\gg$ }  }
\else
\newcommand{\authnote}[3]{}
\fi

\newcommand{\cC}{\mathcal{C}}
\newcommand{\cA}{\mathcal{A}}

\newcommand{\cT}{\mathcal{T}}
\newcommand{\cD}{\mathcal{D}}
\newcommand{\cM}{\mathcal{M}}
\newcommand{\cX}{\mathcal{X}}
\newcommand{\cY}{\mathcal{Y}}
\newcommand{\cV}{\mathcal{V}}
\newcommand{\cE}{\mathcal{E}}
\newcommand{\cB}{\mathcal{B}}
\newcommand{\divmax}{\Delta_{\text{max}}}
\newcommand{\divkl}{\Delta_{\text{KL}}}
\newcommand{\TV}{\mathsf{TV}}
\newcommand{\KL}{\mathsf{KL}}
\newcommand{\Hel}{\mathsf{H}}

\DeclarePairedDelimiter{\abs}{\lvert}{\rvert}

\newcommand{\safe}{\ensuremath{\textsf{safe}}\xspace}
\newcommand{\loosafe}{\ensuremath{\textsf{leave-one-out-safe}}\xspace}
\newcommand{\ssafe}{\ensuremath{\textsf{sharded-safe}}\xspace}
\newcommand{\CPdelta}{\textsf{CP-$\Delta$}\xspace}
\newcommand{\CPk}{\textsf{CP-k}\xspace}
\newcommand{\CPsmooth}{\textsf{smooth-CP-k}\xspace}

\newcommand{\mathLarger}[1]{\mathlarger{\mathlarger{#1}}}

\begin{document}
	
	\begin{toappendix}
\section{Comparison with Differentially Private Prediction} \label{sec:dp}

In Section~\ref{sec:discussbody}, we discussed relations between $k$-near access-freeness ($k$-NAF) and
$\eps$-differential privacy ($\eps$-DP). 
We now make a more detailed comparison with a more closely related
variant of DP, namely
\emph{privacy-preserving prediction}~\citep{dwork2018privacy}, in which  the
aim is to protect the
privacy of a single individual prediction (i.e., outputs) as opposed to the model itself. 
The setting is where a user can only access the model through its predictions.
Here, a mechanism $\cT$ is a randomized mapping that takes as input a
dataset $\cD$ and a prompt $x\in\cX$ and returns a prediction $y\in\cY$.  
For example, $\cT(\cD)(x)$ may be a procedure that
first trains a model $q$ with $\cD$ and then samples $y$ from
$q(\cdot|x)$.  
We say $\cT$ is $\eps$-DP prediction preserving if for every input $x$ and output $y$
\begin{equation}
	e^{-\eps} \Pr[ \cT(\cD')(x)= y ] \leq \Pr[ \cT(\cD)(x)= y ] \leq e^\eps \Pr[ \cT(\cD')(x) = y ] \;. \label{eq:dppred}
\end{equation}
The probability in (\ref{eq:dppred}) is taken over the randomness used
in the mechanism $\cT$, with input $\cD$ and $x$, e.g.  the randomness
is both over the training algorithm used to obtain $q$ and any
randomness used in sampling $y$ from $q(\cdot|x)$. 
\citet{van2020trade} review some DP prediction preserving algorithms,
showing that in some cases these do not provide advantages over
private training of the entire model. 

Three important differences in our definition are: 
(1) our focus is solely on (conditional) generative models $p(\cdot|x)$, and our probabilities are only taken over the distributions induced by these models, while privacy-preserving prediction is concerned with the mechanism's distribution, i.e. the distribution of $\cT(\cD)(x)$ as a function of both $\cD$ and $x$ (where, say to output a label $y$, $\cT$ may require fully retraining on the dataset to learn a deterministic classifier $q(\cdot)$ to use on $x$), 
(2) our probability
comparison is with respect to a given function \safe (a choice left to the user) instead of being defined with respect to $\cT(\cD')$, and (3) our definition's bound is one sided (we only care about an upper bound on the probability of outputting a certain output).
In particular, suppose we have an
algorithm which, upon input $\cD$, returns a model $p$ that is $k$-NAF
with respect to some given function \safe and for $\Delta=\divmax$
(e.g. \CPdelta and \CPk are such algorithms). This implies:
\begin{equation}\label{eq:naf_compare}
p(y|x) \leq 2^k \safe_C(y|x).
\end{equation}
In Section~\ref{sec:twosided-counter} we give a setting where this difference between one sided and two sided is crucial for the feasibility of our algorithms.

Let us observe how we can obtain an
$\eps$-NAF model using an $\eps$-DP prediction preserving
mechanism $\cT$. Define $p(y|x)$ as the probability that
$\cT(\cD)(x)$ outputs $y$, and let us take the \safe
function to be the \loosafe function, where $\cA$ in
Algorithm~\ref{alg:loosafe} is chosen to be $\cT$ itself. Here, the
guarantee in (\ref{eq:dppred}) immediately implies
that $p$ is $\eps$-NAF with respect to \loosafe
and $\Delta=\divmax$. Importantly, note that 
the algorithm $\cA$ used in \loosafe had to be chosen as
$\cT$ in order for this implication to hold.
Conversely, the implication does not
necessarily go in the other direction, i.e. a $k$-NAF model does not necessarily
imply $O(k)$-differentially private prediction, even if we use the function \loosafe.

While this difference between (\ref{eq:naf_compare}) and
(\ref{eq:dppred}) may seem minor at first glance, even here (like for
the general notion of differential privacy), obtaining DP prediction
preserving mechanisms often needs more sophisticated mechanisms while
the condition in (\ref{eq:naf_compare}) is achievable with black box
reductions that do not inject additional randomness.

\subsection{Importance of One Sidedness of NAF} \label{sec:twosided-counter}

We present here a concrete example where the difference between one sided (NAF) and two sided (DP) definitions manifests. Let $\mu$ be a small constant, suppose there is a generative learning algorithm which when trained on a dataset $D\cup\{y_1\}$ yields model $q_1$ and when trained on $D\cup\{y_2\}$ yields $q_2$. Suppose $q_1(y_1)=q_2(y_2)=\mu$ and $q_1(y_2)=q_2(y_1)=\mu^2$ and for all other $y$’s we have $q_1(y)=q_2(y)$. Intuitively, if the model sees the image $y_i$ during training then it outputs $y_i$ with probability $\mu$ which is $1/\mu$ times more likely than the probability of a model outputting $y_i$ which has not seen $y_i$. Depending on the setup this may be a clear case of copyright violation. It is also the case that the underlying learning algorithm only satisfies multiplicative DP with $ \eps=\log(q_1(y_1)/q_2(y_1))=\log(1/ \mu)$ which may be very large. On the other hand the TV distance between the two distributions is only $\approx\mu$ and as \CPdelta for $\Delta = \divmax$ can create a distribution $p$ which only outputs $y_i$ with probability $\approx\mu(1+\mu)$ which is only $1+\mu$ times more likely than the probability of a model outputting $y_i$ which has not seen $y_i$. Note that if our definition was two sided (for $\Delta=\Delta_{max}$) then for all $p$ we have that $\max\{p(y_1)/q_2(y_1),q_1(y_1)/p(y_1)\}\geq1/\sqrt{\mu}$ which possibly much bigger than $1+\mu$.

\end{toappendix}

	\maketitle
	
	\begin{quote}
		\emph{``To constitute an infringement under the Act there must be substantial similarity between the infringing work and the work copyrighted; and that similarity must have been caused by the defendant's having copied the copyright holder's creation.''} --- U.S. 9th Circuit Opinion, Roth Greeting Cards v. United Card Co., 1970. 
	\end{quote}
	
	\begin{quote}
		\emph{``Originality does not signify novelty; a work may be original even though it closely resembles other works, so long as the similarity is fortuitous, not the result of copying. To illustrate, assume that two poets, each ignorant of the other, compose identical poems. Neither work is novel, yet both are original and, hence, copyrightable.''} --- U.S. Supreme Court Opinion, Feist Pubs., Inc. v. Rural Tel. Svc. Co, 1991.
	\end{quote}
	

	\begin{abstract}

		There is a growing concern that learned conditional generative models may output samples that are substantially similar to some copyrighted data $C$ that was in their training set.
		We give a formal definition of \textit{near access-freeness (NAF)} and prove bounds on the probability that a model satisfying this definition outputs a sample similar to $C$, even if $C$ is included in its training set.
		Roughly speaking, a generative model $p$ is \emph{$k$-NAF} if for every potentially copyrighted data $C$, the output of $p$ diverges by at most $k$-bits from the output of a model $q$ that \emph{did not access $C$ at all}. 
		We also give generative model learning algorithms, which efficiently modify the original generative model learning algorithm in a black box manner, that output generative models with strong bounds on the probability of sampling protected content. 
		Furthermore, we provide promising experiments for both language (transformers) and image (diffusion) generative models, showing minimal degradation in output quality while ensuring strong protections against sampling protected content.
		
		
	\end{abstract}
	\section{Introduction}\label{sec:intro}
	
	Generative models for images, text, code, and other domains pose new challenges for ensuring their outputs are protected from copyright infringement.
	Such models are trained on a large corpus of data, where it is often impractical to ensure the training set is 100\% free of copyrighted material. Furthermore, removing copyrighted material from training may also be undesirable.  For example, a human author is free to read and use copyrighted material as inspiration for their work, as long as they do not copy it. Similarly, it may be beneficial to use copyrighted material when training in order to have more effective generative models.
	
	Copyright infringement by generative models can potentially arise in (at least) two manners. First, in the \emph{training phase}, the algorithm could directly access copyrighted material, and the learned model itself 
	could implicitly contain (e.g. coded in its weights) verbatim copies of some of this material.
	The copyright issues arising during training share many similarities with other settings in which algorithms scrape significant amounts of data, including search-engine indexing and digitizing books. 
	Here, the question of what constitutes a copyright infringement is largely a question of  ``fair use.'' 
	This work does not examine these fair use issues that arise in the training phase, and we refer the reader to the several legal precedents in this area~\citep{samuelson2021text}. 
	
	The second notable source of potential infringement is in the \emph{deployment phase}, where a user provides a prompt $x$ to the model to obtain some output $y$. Apriori, we cannot rule out the possibility that $y$ is either a verbatim copy or substantially similar to some copyrighted training data.
	Moreover, unlike search engines, generative models do not keep track of the \emph{provenance} of their outputs. 
	Hence, a user of such an output $y$ (e.g., a software company using generated code, or a designer using a generated image) has no easy way to verify that it does not infringe upon any copyrighted material.
	It is this issue of preventing deployment-time copyright infringement that is the focus of this work. 
	
	\begin{figure*}[t]
		\centering
		\minipage{0.22\textwidth}
		\includegraphics[width=\linewidth]{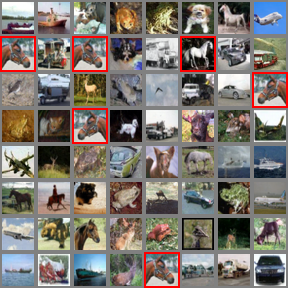}
		\endminipage\hfill
		\minipage{0.22\textwidth}
		\includegraphics[width=\linewidth]{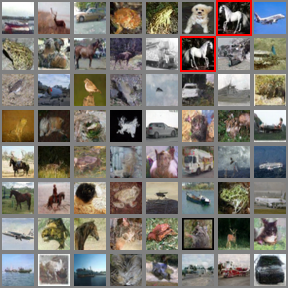}
		\endminipage\hfill
		\minipage{0.22\textwidth}
		\includegraphics[width=\linewidth]{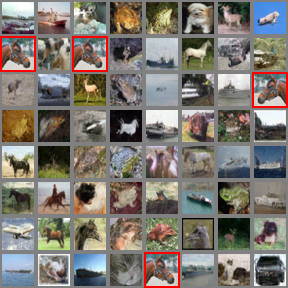}
		\endminipage\hfill
		\minipage{0.22\textwidth}%
		\includegraphics[width=\linewidth]{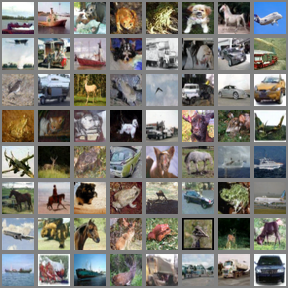}
		\endminipage
		\caption{\textbf{The \CPk Algorithm applied to diffusion models.} {\small  
				The dataset is CIFAR-10 augmented with multiple copies of two images (images close to the augmented images are marked with red boundaries); hypothetically, suppose these two images are copyrighted works. The leftmost image shows generations from a model $p$ that was trained on the full dataset, where we clearly see that $p$ generates the two copyrighted works. 
				Our algorithm starts by splitting this dataset into two
				disjoint datasets, making sure that copyrighted images are split into two different shards; for
				illustrative purposes, we do not deduplicate the dataset. 
				The
				procedure then trains two models $q_1,q_2$ on these disjoint
				shards. The middle two figures show samples from the models $q_1, q_2$, again clearly showing memorization.
				However, note that $q_1$ does
				not generate one of the copyrighted images and and $q_2$ does not generate the
				other copyrighted image (as these were not in their respective datasets).  Our
				algorithm \CPk then uses $q_1,q_2$, along with the original model $p$, to construct a model $p_k$ which has strong copyright protection guarantees. The last image is the outputs of $p_k$, showing it is highly unlikely to output either of the copyrighted images, even though each of $q_1,q_2$ and $p$ has
				memorized some of these images. See Section~\ref{sec:empirical} for more details (and for a discussion with regards to our displayed model generations having used the same noise on the diffusion paths).}
		}
		\label{fig:diffusion-intro-fig}
	\end{figure*}

	\paragraph{Our contributions.}
	We give a formal definition --- ``near-access freeness'' ---  bounding the extent to which a learned generative model's output can be substantially influenced by a particular piece of copyrighted data that the model was trained on. 
	We also give a procedure that transforms (under certain assumptions) any generative model learning algorithm $\cA$ into an algorithm $\cA_k$, which protects against violations under our definition. 
	In particular, the model output by $\cA_k$ will (1) be at most
	$k$-bits far from a ``safe'' model (which is not committing copyright
	infringement), and (2) will have performance reasonably close to the model output by the original algorithm $\cA$ (in a quantifiable sense, based on properties of $\cA$).  Our algorithms have a relatively modest multiplicative overhead in training and inference compared to $\cA$ itself. 

	We also show promising experiments on language and image generative models, demonstrating that our modified model does not degrade significantly in quality (and in fact, it may even improve in some cases). See Figure~\ref{fig:diffusion-intro-fig} for one example and Section~\ref{sec:empirical} for more details. 
	
	
	Our definition satisfies a few notable properties:
	\begin{description}
		\item[Separation of access and similarity:] Demonstrating a copyright infringement consists of showing both \emph{access} to the copyrighted material and \emph{substantial similarity} of the output to the material. Our definition separates these two aspects, considering an abstract access function (that can have several different practical realizations) and a quantitative measure of similarity.
		
		\item[Information-theoretic measures:]  Our framework defines similarity on the \emph{probability distributions} of generative models rather than on particular outputs themselves. This enables using information-theoretic similarity measures, rather than being restricted to superficial notions of similarity such as Hamming or edit distance. 
		
		\item[Similarity relative to a ``safe'' baseline:] 
		We measure the degree of similarity between our model's output and some copyrighted data $C$, by comparing the likelihood of our model generating $y$ to that of some \emph{safe} model, which was trained without access to $C$. This matches copyright law which (unlike patents) allows for accidental or ``fortuitous'' similarity (see quote above from Feist vs. Rural). In this sense, our definition bears some similarities to \emph{differential privacy}, though there are significant differences as well; see  Section~\ref{sec:discussbody}.
	\end{description}
	
	\paragraph{Organization.} 
	Section~\ref{sec:definition} presents our definition and discusses
	some motivations and implications;  Section~\ref{sec:algorithms}
	provides provably efficient algorithms that modify a baseline training
	algorithm $\cA$ into a version that is protected, under our
	definition; Section~\ref{sec:empirical} provides a brief experimental
	validation;
	and Section~\ref{sec:prior} and~\ref{sec:discussion} present prior work and our concluding remarks. All proofs not in the main body are deferred to the appendix.
	
	\paragraph{Note / Disclaimer.} Generative models raise many legal and
	ethical issues. This paper focuses on copyright infringement by the
	outputs of generative models, which
	is only one of these issues. The concepts and tools we provide do not
	address issues related to other forms of intellectual property,
	including \emph{privacy},  \emph{trademarks}, \emph{patents}, or \emph{fair
		use}. Moreover,  
	our work does not (and cannot) guarantee the absence of copyright
	infringement in all settings. However, we do hope it provides helpful tools and concepts that can be used by model creators and users, lawyers, and courts to reduce the task of determining if some types of infringements have occurred to well-defined, quantitative questions.

	\section{Near Access-Free Generative Modeling} \label{sec:definition}
	
	Our setting is as follows:
	we assume there is an algorithm $\cA$ which takes as
	input a dataset $\cD=\{z_1, \ldots z_N\}$ and returns a
	conditional generative model $p(\cdot|\cdot)\in\cM$, where
	$\cM$ is the space of conditional generative models.  Here,
	the conditional generative model $p$ can take some prompt
	$x\in \cX$ as input and then outputs $y\in\cY$ with
	probability $p(y|x)$. We can think of $x\in\cX$, $y\in\cY$, and $z\in\cD$
	as program snippets, sequences of text,  images, etc. 
	
	Some training samples in our dataset $\cD$ may contain copyrighted material. We let $\cC$ be the
	set of copyrighted material contained in $\cD$, e.g. $C \in \cC$ may
	be a snippet of code, text, or artwork that is contained in one or more training samples in $\cD$.
	We are concerned that our algorithm may return a model $p$ that samples copyrighted material from
	$\cC$ (or material substantially similar to that in $\cC$) with non-trivial probability. That is, for
	$p=\cA(\cD)$, the concern is that for some prompt $x$ and copyrighted material
	$C\in\cC$, it holds that $y \sim p(\cdot|x)$ will be similar to  $C$ with non-trivial probability. Our goal is to devise a procedure where this is not the case. 
	
	
	\subsection{$k$-Near Access-Freeness}\label{sec:knaf}
	Under the laws of the U.S. and many other countries, to establish a copyright infringement, a plaintiff must  prove that \textbf{(1)} ``the defendant had \emph{access} to the plaintiff’s copyrighted work'', \textbf{(2)} ``there are \emph{substantial similarities} between the defendant’s work and original elements of the plaintiff’s work.''\footnote{See U.S. Courts for the 9th circuits, Model Civil Jury instructions, \href{https://www.ce9.uscourts.gov/jury-instructions/node/274}{Section 17.17 Copying -- Access and Substantial Similarity}; emphases ours. Even if the access and substantial similarity tests pass, it may still be considered ``fair use'', but since the fair use condition is more application-dependent, we do not consider it here, making our definition more conservative.}  Our definition is modeled around these two components of \emph{access} and \emph{substantial similarity}. 
	
	Informally, 
	our goal is to provide a model $p$ such that, for any prompt $x$ and copyrighted element $C\in\cC$, 
	the distribution  $p(\cdot|x)$ is within $k$-bits of information (measured under some divergence measure) to a
	``safe'' generative model, which was trained without access to
	$C$. We now formalize this notion.
	We use the abstraction of a function \safe that maps 
	a datapoint $C\in\cC$ into a generative model $\safe(C)\in\cM$ that is assumed to
	have been trained without any \emph{access} to $C$. (For notational
	convenience, we sometimes overload notation by denoting $\safe(C)$ as $\safe_C$.)
	For example, the \loosafe function, shown in
	Algorithm~\ref{alg:loosafe}, is one such example; in this construction, $\cD_{-C}$ refers to the dataset
	where \emph{all} datapoints that access $C$ have been removed.
	
	Since $\safe(C)$ is a generative model that
	was learned without access to $C$, in many realistic scenarios the
	probability that $\safe_C(\cdot|x)$ generates material that is similar
	to $C$ itself will be \emph{exponentially small} in the length of $C$
	(though see Section~\ref{sec:discussbody} for when this may not be the
	case).
	Moreover, even if this unlikely event happened, this generation
	can be said to be fortuitous (see the quote from Feist vs. Rural above the abstract).
	
	
	\begin{algorithm}
		\caption{\textsf{leave-one-out-safe}} \label{alg:loosafe}
		\centering
		\begin{minipage}{4in}
			\begin{algorithmic}
				\BeginBox
				\Procedure{Leave-One-Out-Safe}{}
				\State \textbf{Input:}  Dataset $\cD$
				\State \textbf{Output:} the following mapping from $\cC$ into $\cM$, where
				\[
				\loosafe(C) = \cA(\cD_{-C})
				\]
				\EndProcedure
				\EndBox  
			\end{algorithmic}
		\end{minipage}
	\end{algorithm}
	
	We now introduce our main criterion for copyright protection, which combines the notion of access, as provided through some prespecified function $\safe$, with the
	notion of substantial similarity.
	
	\begin{definition}[$k$-Near Access-Free] \label{def:bounded}
		Let $\cC$ a set of datapoints; let \textsf{safe}$:\cC \rightarrow \cM$; 
		and let $\Delta$ be a divergence measure between
		distributions.  We say that a generative model $p$ is \emph{$k_x$-near
			access-free ($k_x$-NAF)} on prompt $x\in\cX$ with respect to $\cC$, 
		$\safe$, and $\Delta$ if for every $C \in \cC$,
		\begin{equation} 
			\Delta\Bigl( \, p(\cdot|x) \;\mathLarger{\|}\; \safe_C(\cdot|x) \, \Bigr) \leq k_x \;. \label{eq:prob-bound}
		\end{equation}
		We say $p$ is $k$-NAF if the above holds for all $x\in\cX$ with $k_x\leq k$.
	\end{definition}
	When clear from context, we drop the $\cC$, $\safe$, and $\Delta$ dependence and simply say $p$ is $k_x$-NAF  on input $x$.
	
	Definition~\ref{def:bounded} reduces the task of determining a copyright infringement to (1) a \emph{quantitative} question of the acceptable value of $k$, and (2) a \emph{qualitative} question of providing a \safe function that appropriately satisfies a no access condition. Both can be application-dependent: the number of bits that constitute copyrightable content differs between, e.g., poems and images, and the \safe function could also differ based on application. 
	However, with respect to an acceptable $k$ and a given function $\safe$, a model satisfying Definition~\ref{def:bounded} provides a rigorous guarantee of no substantive similarity.
	
	\paragraph{Choices for the divergence measure.}
	Our default choices will be either the \emph{maximum KL divergence} $\divmax$, also known as the Rényi divergence of order infinity, or the \emph{KL divergence} $\divkl$. For two distributions $\rho,\mu$,  $\divmax(\rho\|\mu) = \max_{y\in \mathrm{Supp}(\rho)} \log \tfrac{\rho(y)}{\mu(y)}$ and $\divkl(\rho \| \mu) = \E_{y\sim \rho}\log\left(\tfrac{\rho(y)}{\mu(y)}\right)$.~\footnote{Throughout this paper, $\log$ is with base $2$.}
	Let us now formalize our motivation for using these measures, starting with $\divmax$. 
	Plugging $\Delta=\divmax$ in Definition~\ref{def:bounded} rules out simple copying and even copying substantial components of the copyright text, which we formalize with the following lemma:

	\begin{lemma}[Event bound, max-KL] \label{lem:nocopying} Suppose
		model $p$ is $k_x$-NAF on prompt $x$ with respect to
		$\cC,\safe, \Delta=\divmax$. Then for any $C\in\cC$ and any event
		$\cE$,
		\[
		p(\cE|x) \leq 2^{k_x} \cdot \safe_C(\cE|x).
		\]
	\end{lemma}
	The proof directly follows from the definition of $\divmax$. 
	\begin{proof}
		By definition,  $\divmax(p(\cdot | x) \| \safe_C(\cdot|x) ) \leq k_x$ implies that  for every $y$, $p(y|x) \leq 2^{k_X} \safe_C(y|x)$.
		The result follows from summing over all $y\in \cE$.
	\end{proof}
	
	For some copyrighted text $C\in \cC$, let $V_C$ be the event that the output is substantially similar to $C$. 
	Lemma~\ref{lem:nocopying} implies that
	\begin{equation*}
		\underbrace{\color{red} p(V_C|x)}_{\text{\rm \color{red} probability of violation}} \leq \;\;\; 2^{k_x} \;\;\; \cdot \; \;\; \!\!\!\!\!\!\!\!\!\!\!\!\!\!\!\! \underbrace{\color{blue} \safe_C(V_C|x).}_{\text{\rm \color{blue} probability of violation with access-free model}}   
	\end{equation*}
	As we expect the probability of $V_C$ under $\safe_C(\cdot|x)$ 
	to be exponentially
	small in the length of the output (since  $\safe_C$ was trained
	without access to $C$, though see Section~\ref{sec:discussbody}), this would then imply that $p$ itself has a small violation probability.
	
	Our motivation for considering $\divkl$ is that it satisfies a similar bound as $\divmax$, under slightly stronger assumptions. Let us say that a random variable $X$ is \emph{$(\eps,\delta)$-concentrated} if $\Pr[ X\notin (1\pm \eps)\E[X]
	] \leq \delta$. Analogous to the previous lemma, we have:
	\begin{lemmarep}[Event bound, KL concentrated] \label{lem:nocopyingklconc}
		Suppose model $p$ is $k_x$-NAF on prompt $x$ with respect to
		$\cC,\safe, \Delta=\divkl$, and suppose the random variable $Y_x= \log \tfrac{p(y|x)}{\safe_C(y|x)}$ (with $y\sim p(\cdot|x)$) is
		$(\eps_x,\delta_x)$-concentrated. Then, for any $C\in\cC$ and any event
		$\cE$,
		\[
		p(\cE|x) \leq 2^{(1+\eps_x) k_x} \cdot \safe_C(\cE|x) + \delta_x.
		\]
	\end{lemmarep}
	\begin{proof}
		For every prompt $x$, let $Y_x$ be as above, and define $\cB=\cB_x$ to be the event $Y_x \not\in (1\pm \eps_x)\E[Y_x]$. Under our assumptions $\E[Y_x] = \divkl(p(\cdot|x),\safe_C(\cdot|x)) \leq k_x$ and (due to concentration) $\Pr[ \cB] \leq \delta_x$. Now for every event $\cE$, we can write $p(\cE|x) = p(\cE \cap \overline{\cB} | x) + p(\cE \cap \cB | x)$.
		The first term is  $\sum_{y \in \cE \cap \overline{\cB}} p(y|x) \leq \sum_{y\in \cE \cap \overline{\cB}}2^{(1+\eps_x)k_x} \safe_C(y|x)$ since for every $y\in \overline{\cB}$, $\log \tfrac{p(y|x)}{\safe_C(y|x)} \leq (1+\eps_x)k_x$. The second term is bounded by $p(\cB|x) \leq \delta_x$.
		So we get 
		\begin{multline*}
			p(\cE|x) \leq \sum_{y\in \cE \cap \overline{\cB}}2^{(1+\eps_x)k_x} \safe_C(y|x) + \delta_x 
			= 2^{(1+\eps_x)k_x}\safe_C(\cE \cap \overline{\cB}|x) + \delta \leq 2^{(1+\eps_x)k_x}\safe_C(\cE|x) + \delta_x
		\end{multline*}
	\end{proof}
	
	The relation between the bounds of Lemma~\ref{lem:nocopying} and Lemma~\ref{lem:nocopyingklconc}  is somewhat analogous to the relation between $\eps$-differential privacy and  $(\eps,\delta)$-differential privacy \citep{DPsurvey}.
	

	\subsection{Further Discussion of the Definition} \label{sec:discussbody}
	
	\paragraph{Is $\safe_C(y|x)$ exponentially small?} In many settings,
	we expect $\safe_C(V_C|x)$ to be small due to $V_C$ corresponding to a ``monkeys on typewriter'' event, whereby a
	process with no access to $C$ produced a copy of $C$ by accident. 
	However, consider the prompt $x =$  \texttt{"print the following text:$C$"}, where the text in $C$ itself has been inserted into the prompt.
	In such a case, even the ``safe'' model $\safe_C$ will output $C$ on $x$ with high probability and so $\safe_C(V_C|x)$ will \emph{not} be exponentially small.
	Yet our definition can still be satisfied (and in particular it vacuously holds that $p(V_C|x)$ is not much larger than $\safe_C(V_c|X)$).
	We view this as a reasonable outcome because the behavior of $p$ is similar to that of a procedure which had \emph{no access} to $C$. Crudely, an analogy would be to copy-paste a copyrighted text into a word processor, which would not be considered a copyright violation due to the word processor software.
	
	We view subtle cases like this as a strength of the framework, as our
	definition serves as means to quantitatively discuss such questions.

	\paragraph{Comparison with Differential Privacy.} 
	At first look, it may seem that \emph{near access-freeness (NAF)} is equivalent to the well-known notion of \emph{differential privacy  (DP)} \citep{DworkMNS06}, with the parameter $k$ playing the role of
	$\eps$. But in fact, there are crucial differences between the two, which we now discuss.
	
	First, the \emph{goals} of privacy and copyright protection, while
	related, differ in important ways. Privacy is focused on an
	\emph{individual} and the attributes of that individual while
	copyright protection is only for a specific piece of work.
	Moreover, copyright only protects the specific expressions of that work, and not the ideas present in it.%
	\footnote{This is known as the \emph{idea/expression dichotomy} whereas copyright only protects a particular \emph{expression} of an idea rather than the idea itself~\citep{samuelson2007copyright}. The Copyright Act of 1976 asserts  that \emph{``In no case does copyright protection for an original work of authorship extend to any idea, procedure, process, system, method of operation, concept, principle, or discovery, regardless of the form in which it is described, explained, illustrated, or embodied in such work''} and the \href{https://www.govinfo.gov/content/pkg/USCODE-2019-title17/html/USCODE-2019-title17.htm}{legislative history} clarifies that \emph{``copyright does not preclude others from using the ideas or information revealed by the author's work.''}}  
	For example, if the output of a machine-learning procedure leaks a particular piece of information (e.g., medical diagnosis) about an individual, then this
	is a privacy violation no matter how the information is expressed, while
	the form of an expression is crucial in the context of copyright.
	This difference is also translated into quantitative terms:
	if any particular generative output leaks even a few bits about a training sample, this could still
	be a significant privacy violation.
	In contrast, a few bits of leakage are unlikely to constitute a copyright violation since copyright requires a minimum amount of information content.\footnote{Indeed, the U.S. Copyright Office states \citep{circular} that ``Words and short phrases, such as names, titles, and slogans, are uncopyrightable because they contain an insufficient amount of authorship.''}
	More generally, privacy requires that the output of a mechanism does not reveal whether or not an individual's data was in the database.
	For copyright protection, we only need to ensure that no particular output is substantially similar to a copyrighted work to which it had access, and it is explicitly allowed for the model to use ``ideas or information'' revealed by the copyrighted works it was trained on.

	Given the above differences, it is not surprising that the algorithms
	to ensure privacy and copyright protection would differ and also
	exhibit different performance tradeoffs.  
	This is indeed the case.
	To elaborate more, we recall the definition of differential privacy.
	Let $\cT$ be a mechanism that maps datasets to generative models. 
	We say that $\cT$ is \emph{differentially private (DP)} if for every datasets $\cD$ and $\cD'$ that differ by at most one point, and every model $p\in\cM$ in the support of $\cT$,
	\begin{equation}
		e^{-\eps} \Pr[ \cT(\cD')= p] \leq \Pr[ \cT(\cD)= p ] \leq e^\eps \Pr[ \cT(\cD')= p] \;. \label{eq:dp}
	\end{equation}
	The probability in (\ref{eq:dp}) is taken over the randomness used in
	the mechanism $\cT$, with input $\cD$.  The Near-Access-Free condition
	is not explicitly concerned with the model itself, but only with
	outputs from the model.  For example, a neural model whose weights
	encode the entire training set would completely violate differential
	privacy, but, so long as the model never generates particular outputs that are similar to
	the copyrighted data, it may very well not violate copyright.  
	In that sense, our definition is closer to \emph{privacy-preserving
		prediction}~\citep{dwork2018privacy}, which aims to protect the
	privacy of individual predictions (i.e., outputs) as opposed to
	protecting the model itself. Even here there are important
	technical distinctions, which we discuss in Appendix~\ref{sec:dp}.

	It is worth noting that these differences have important algorithmic
	implications. Achieving privacy-preserving mechanisms
	often requires the use of carefully constructed mechanisms (which inject
	additional randomness into the models and/or training).  In contrast, as our main results show, near-access-freeness 
	is achievable with black-box reductions, requiring only some base learning algorithm $\cA$ (and no additional randomness).
	Also, a series of
	papers have been exploring the effectiveness of privacy-preserving
	methods using neural models, which suggest either better features are
	needed or that more sophisticated approaches are required
	(e.g. \cite{tramer2021differentially,li2021large,ghazi2021deep}).  Of
	course differential privacy
	provides stronger privacy guarantees while
	near-access-freeness is only designed to protect against copyright
	infringement.

	\paragraph{The $\safe$ function in practice.} There can be a number of
	different ways to define the $\safe$ function in practice. The
	``leave-one-out'' example is one, but it requires the training of $|\cC|$ different models. We
	describe a far more efficient implementation in
	Section~\ref{sec:algorithms}. In both cases, it is important that when we omit a datapoint $x$ it does not share copyrighted content with many other datapoints that were included in the training set. If we assume that datapoints that share the same copyrighted content are near-duplicates we could achieve this by \emph{deduplication}. But in general this may not be the case, in such situations we could cluster the dataset by content or by metadata such as authorship so that all works which are close (and hence possibly share copyrighted content) are omitted together. If we can ensure this then we can use our implementations as is. In practice, such processes will be likely approximate, still we think that they should be sufficient for most copyrighted works. For simplicity, we will assume in Sections~\ref{sec:cpdelta} and~\ref{sec:cpk} all copyrighted works occur at most one datapoint and we will relax our assumption to $m$ datapoints in Section~\ref{sec:largerm}.
	While we will implement \safe by partitioning the dataset $\cD$ into two (or more, see Section~\ref{sec:largerm}) parts, there may be other ways to ensure safety. 
	For example, the output of $\safe$ might be a model trained  ``golden
	dataset'' which is much smaller but was carefully scrutinized to
	ensure that all material in it is not copyrighted or properly
	licensed. Another way to ensure that a model is safe for $C$ is to train it only on data that was generated before $C$'s creation. 
	
	\paragraph{What is $\cC$?} In our discussions, we refer to $C\in\cC$
	abstractly as a ``piece of copyrighted data'', but do not specify it
	in more detail. For example, in an image generative model, does $C$
	correspond to a single artwork, or the full collected arts of some
	artists? The answer is the former. The reason is that if a generative
	model generates data that is influenced by the full collected artworks
	of X, but not by any single piece, then it is not considered a
	copyright violation. This is due to that it is not possible to copyright style or ideas, only a specific expression. Hence, we think of $C$ as a piece of content that is of a similar scale to the outputs of the model.
	
	\paragraph{Comparison with law.} 
	As discussed earlier to show a copyright violation has occurred the plaintiff must prove that``there are \emph{substantial similarities} between the defendant’s work and original elements of the plaintiff’s work'' (assuming \emph{access}). It's negation would be to show that defendant’s work is not substantially similar to the original elements of the plaintiff’s work. Our approach would instead correspond to showing that the defendant’s work is close to a work which was produced without access to the plaintiff’s work. While we think this is a stronger guarantee, to what extent the courts (or lawmakers / regulators) will embrace it is an open question.

	\paragraph{Other choices for divergence measure.} The divergence
	measure $\Delta$ need not be Max-KL or KL. Other choices include the following:
	\begin{description}
		
		\item[Earthmover metrics.] In many settings, whether text or images, there is a natural context-dependent geometry over images, and so there is a pointwise divergence measure $\delta(y\|y')$ which measures some notion of distance (i.e., a quasi metric) between $y$ and $y'$ in the support of the distributions.  The function $\delta$ could be simply the Hamming distance, but could also take into account context-specific measures such as edit distance or semantic similarity in natural language text, syntactically equivalent transformations for programs, or visual transformations such as cropping, rotating, and filtering in the context of images. The \emph{earthmover distance} of $\rho,\mu$ is the minimum of $\E_{(y,y')\sim \tau} [\delta(y\|y')]$ over all \emph{couplings} $\tau$ of $\rho,\mu$ (i.e., distributions whose marginals are $\rho$ and $\mu$ respectively). 
		
		\item[Combination metrics.] It is possible to combine the two aspects above, and define $\Delta(\rho\|\mu)$ as the minimum of $\divmax(\rho'\|\mu)$ over all $\rho'$ that are of at most some earthmover distance $D$ to $\rho$. This can combine the advantages of both metrics. Of course, the acceptable value for $D$ would be application dependent.

		\item[Metrics for long sequences.] If the models generate very long sequences of information (e.g., several pages of text, or a full program), then it may be appropriate to consider definitions that look at \emph{subsequences} of the reference and safe model. For example, it may be appropriate for 10 pages of a generated output to include 100 words from some copyrighted data, as long these are ``spread out'' and not part of (say) a 200-word subsequence.
		
	\end{description}
	
	The right choice of the metric will be context dependent. While
	$\divmax$ is very stringent and gives a hard bound in terms of entropy
	on the amount of non-accidental copying, it might be too stringent in
	some applications, ruling out models that are arguably still safe.

	\begin{algorithm}
		\caption{\ssafe} \label{alg:ssafe}
		\centering
		\begin{minipage}{4in}
			\begin{algorithmic} 
				\BeginBox
				\Procedure{Sharded Safe}{} 
				\State \textbf{Input:}  Dataset $\cD$
				\State \textbf{Shard $\cD$:} Partition $\cD$ into
				two datasets $\cD_1$ and $\cD_2$.
				\State \textbf{Learning $\cD$:} Set
				$
				q_1 = \cA(\cD_1), \, q_2 = \cA(\cD_2)
				$
				\State \textbf{Return:}  $q_1$, $q_2$, and the function 
				\[
				\ssafe(C):=q_i,\, \textrm{where } C\notin \cD_i
				\]
				\EndProcedure
				\EndBox
			\end{algorithmic}
		\end{minipage}
	\end{algorithm}

	\section{Algorithms for Copyright Protection} \label{sec:algorithms}
	We now show there exist algorithms for learning a conditional generative model $p$ that can satisfy
	the $k$-NAF condition, for reasonable choices of $k$.  
	For the intuition of our construction, note that for large datasets we may expect that
	$\loosafe(C)\approx\loosafe(C')$, for all $C,C'\in \cC$. The
	algorithmic challenge would then be to find a model $p$ which agrees, under $\Delta$, with $\loosafe(C)$
	for all choices of $C$, and, thus, the model $p$ itself should be
	close to a model that has been trained without access to any $C\in
	\cC$.  This may be computationally difficult because $\loosafe(C)$ is
	one of $|\cC|$ different models.
	
	Now let us see how to make this approach more tractable.
	For simplicity, in Section~\ref{sec:cpdelta}, we assume that each copyrighted piece of data $C$ appears in at most a single datapoint in the dataset.
	While in some settings this can be achieved via deduplication, our
	constructions extend naturally to the case that each copyrighted work
	appears in no more than $m>1$ points (see Section~\ref{sec:largerm}).
	Proceeding under this assumption, we use the function  \ssafe (see Algorithm~\ref{alg:ssafe}). 
	Given a dataset of $N$ points, \ssafe trains two models, each on $N/2$ disjoint points.
	In contrast to \loosafe, we have that, for all $C\in\cC$,  $\ssafe(C)$ is only one of two models, either $q_1$ or $q_2$, corresponding to the model which was not trained on $C$. 
	Our algorithmic challenge is now to find a $p$ which approximately but simultaneously agrees, under $\Delta$, with both $q_1$ and $q_2$.

	Section~\ref{sec:cpdelta} starts by providing algorithms which satisfy the $k$-NAF property
	for both the $\divmax$ and $\divkl$ divergences with respect to the $\ssafe$ function.
	In both cases, the quantity $k$ will be controlled by a distance between the distributions $q_1$ and
	$q_2$; the relevant distance will be the total variation distance when $\Delta=\divmax$ and 
	the squared Hellinger distance when $\Delta=\divkl$. Also, in both cases, we only need the distributions to
	have very mild overlap (i.e., distance slightly bounded away from $1$)
	to ensure a meaningful bound on $k$. Section~\ref{sec:cpk} then considers a more practical, black box
	approach to achieving copyright protection.  Section~\ref{sec:largerm}
	extends our \ssafe construction so that it is applicable if
	each $C\in\cC$ possibly appears in up to some $m>1$ points in $\cD$.
	
	\subsection{The \textsf{Copy-Protection-$\Delta$} Algorithm}\label{sec:cpdelta}
	
	This section assumes  each copyrighted piece of data $C$
	appears in at most a single datapoint in the dataset. The \CPdelta Algorithm is presented in Algorithm~\ref{alg:CP}. 
	Here, $\Delta$ is chosen to be either the $\divmax$ or $\divkl$ divergences.
	Recall that the \emph{total variation distance} between distributions $p$ and
	$q$ is defined as $\TV(p,q) = \tfrac{1}{2}\sum_y |p(y)-q(y)|$ and the \emph{Hellinger squared distance}
	is defined as $\Hel^2(p,q) = 1- \sum_y \sqrt{p(y)q(y)}$. 
	
	Our main result for \CPdelta follows: 
	
	\begin{theorem}[\CPdelta]
		\label{thm:cpalg}
		Let $p$ be the model returned by \CPdelta, and $q_1$ and $q_2$ be the models returned by \ssafe.
		We have that $p$ is $k_x$-NAF with respect to $\cC$, $\ssafe$, and $\Delta$, where\footnote{The factor of $2$ in the case of $\Delta=\divkl$ is not inherent, and can be eliminated in several cases. Whenever that is the case, the bound on $k_x$ is better since for every two distributions, the squared Hellinger distance is upper bounded by the total variation distance.}
		\[
		k_x \leq \begin{cases}
			-  \log \left(  1-\TV\bigl( q_1(\cdot|x)  ,  q_2(\cdot|x) \bigr)\right)   & \textrm{if } \Delta=\divmax  \\
			- 2\cdot \log \left(  1-\Hel^2\bigl( q_1(\cdot|x)  ,  q_2(\cdot|x) \bigr)   \right)  & \textrm{if } \Delta=\divkl.
		\end{cases}
		\]
	\end{theorem}
	
	By Lemma~\ref{lem:nocopying}, for the case of $\divmax$,  we have that
	for all $C\in\cC$ and events $\cE$, 
	\begin{equation}\label{eq:nocopying_divmax}
		p(\cE|x) \, \leq \, \frac{\ssafe_C(\cE|x)}{1-\TV\bigl( q_1(\cdot|x)  ,  q_2(\cdot|x) \bigr)}. 
	\end{equation}
	In other words, provided $q_1$ and $q_2$ have total variation distance  only non-trivially
	bounded away from $1$ and if the probability of a fortuitous copy is
	small, then $p$
	will also copy with only a small probability. 
	
	\begin{algorithm}
		\caption{\CPdelta} \label{alg:CP}
		\centering
		\begin{minipage}{4.2in}
			\begin{algorithmic} 
				\BeginBox
				\Procedure{\CPdelta: \textnormal{Copy Protection w.r.t. divergence $\Delta$}}{} 
				\State \textbf{Input:}  Dataset $\cD$, and
				divergence $\Delta \in \{ \divmax, \divkl\}$.
				\State \textbf{Learning:} Call $\ssafe(\cD)$ to
				obtain $q_1$ and
				$q_2$.
				\State \textbf{Return:} the model $p$, where:
				\[
				p(y|x) = \begin{cases}
					\frac{\min\{q_1(y|x), q_2(y|x)\}}{Z(x)}    & \textrm{if } \Delta=\divmax  \\
					\frac{\sqrt{q_1(y|x) \cdot q_2(y|x)} }{Z(x)} & \textrm{if } \Delta=\divkl.
				\end{cases}
				\]
				\EndProcedure
				\EndBox
			\end{algorithmic}
		\end{minipage}
	\end{algorithm}

\begin{figure}
		\centering
		\includegraphics[width=4.5in]{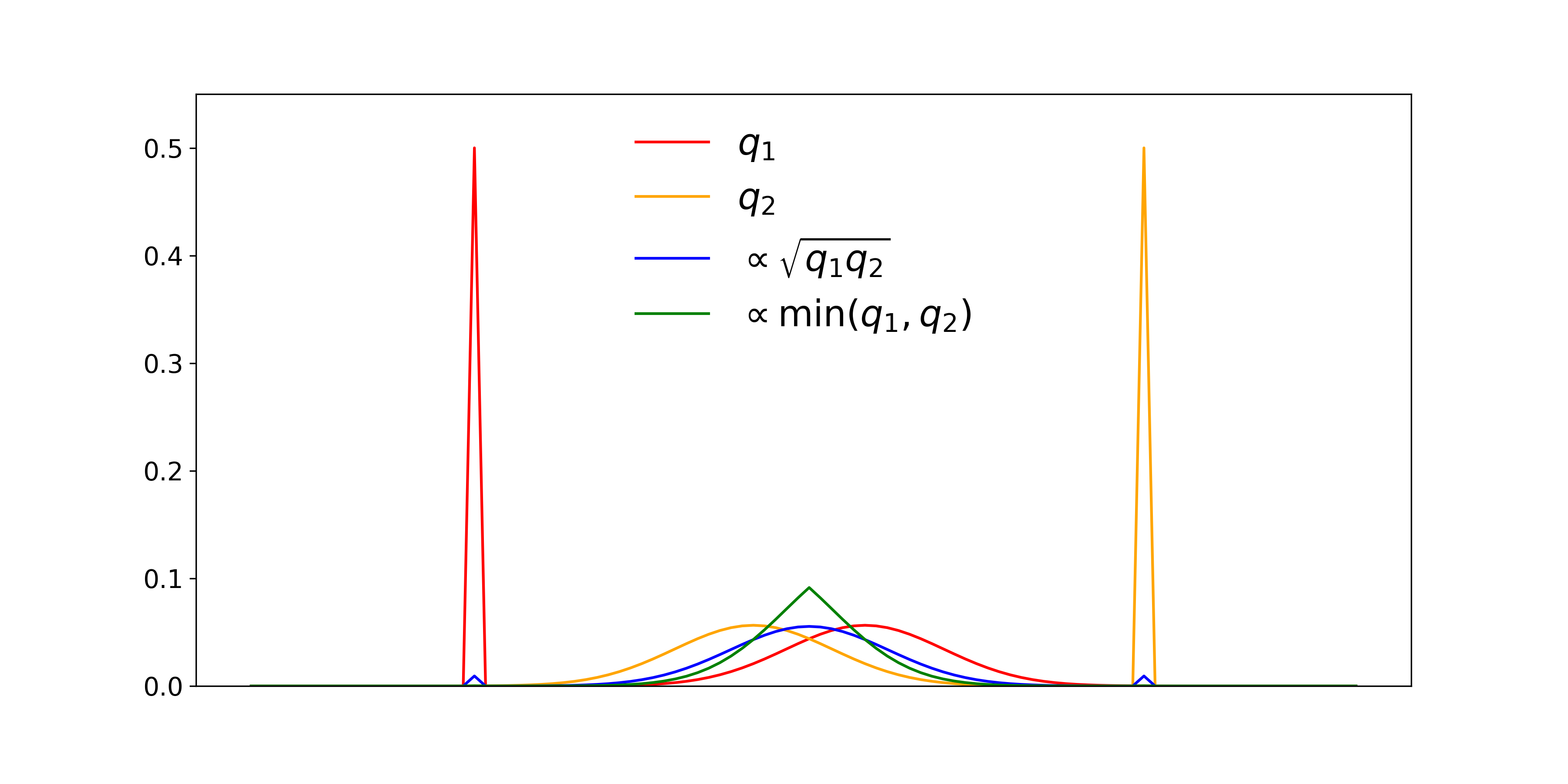}
		\caption{\small 
			Best viewed in color. Applying our transformations to two
			distributions $q_1,q_2$ such that each has a 50\%  chance of
			outputting a different fixed element (the ``spike''), and the remaining distributions have non-trivial overlap. We can interpret each model's ``spike'' as the probability that the model is outputting a ``memorized'' sample from its training set.  Both the distribution proportional to $\min\{q_1,q_2\}$ and the one proportional to $\sqrt{q_1q_2}$ (corresponding to $\CPdelta_{\text{max}}$ and $\CPdelta_{\text{KL}}$ respectively)  significantly suppress the probability of the fixed element while approximately preserving the other probabilities. Also see Example~\ref{ex:cool_example}.}
		\label{fig:exampledists}
	\end{figure}

	Before providing the proof, let us provide an illustrative
	example, which also shows our bound is tight. See Figure~\ref{fig:exampledists} for another example.

	\begin{example}\label{ex:cool_example}
		Consider the promptless case (i.e. $\cX=\emptyset$), where
		there are two (distinct) copyright
		elements, $\cC=\{C_1,C_2\}$, appearing only once each in our dataset
		$\cD$; $\cD$ may contain other training datapoints. Let $\cD_1$ and $\cD_2$ be the dataset
		split, where $\cD_i$ contains $C_i$ for $i\in\{1,2\}$ (and $\cD_i$
		does not contain $\cC_{-i}$). Let $q_i=\cA(\cD_i)$ be
		the model returned by our algorithm on $\cD_i$. Suppose that $q_i(y) = 0.5\cdot I(y=C_i) + 0.5\cdot q(y)$, where we
		interpret $q$ to be the common part learned by both $\cA(\cD_1)$ and
		$\cA(\cD_2)$. As such, we expect $q(\cC)$ to be extremely small, and for simplicity, we assume $q(C_1)=q(C_2)=0$.
		Each of the models $q_1,q_2$ outputs a copyrighted text with probability $1/2$, and yet one can verify that both the distribution proportional to $\min \{q_1,q_2\}$ and the one proportional to $\sqrt{q_1 q_2}$ will simply be $q$.  Hence, the output model of \CPdelta will \emph{never} output $C_1,C_2$.
		For every $y$ in the support of $q$ and for $i\in \{1,2\}$, $q(y)/q_i(y)=2$ and so $\divmax(q(\cdot),q_i(\cdot))=\divkl(q(\cdot),q_i(\cdot))=\log 2$.
		On the other hand, it is easy to see that $\TV(q_1,q_2)=\Hel^2(q_1,q_2)=1/2$.
		Hence, the bound $\divmax(q,p_i) \leq -\log (1-\TV(q_1,q_2))$ is tight and the bound $\divkl(q,p_i) \leq -2\log (1-\Hel^2(q_1,q_2))$ is loose by a factor of two.
		A more general case of how both algorithms apply to two ``spiked'' distributions is illustrated in Figure~\ref{fig:exampledists}. 
		
		
	\end{example}
	
	
	\paragraph{Proof of Theorem~\ref{thm:cpalg}.}
	We start by relating $k_x$, in each case, to the corresponding partition function $Z(x)$. First, for $\Delta=\divmax$, observe that, by construction, $p(y|x) \leq q_i(y|x)/Z(x)$ 
	for all $y\in\cY$, $i\in\{1,2\}$. Hence, $\log(p(y|x)/q_i(y|x))
	\leq \log(1/Z(x))$, and this directly implies that $p$ is
	$\log(1/Z(x))$-NAF. For $\Delta=\divkl$, we have that:
	\begin{align*}
		k_x &= \max_{i \in \{1, 2\}} \KL(p(\cdot|x), q_i(\cdot|x))\\
		&\leq \KL(p(\cdot|x)\|q_1(\cdot|x)) + \KL(p(\cdot|x)\|q_2(\cdot|x))\\ 
		&= \E_{y\sim p(\cdot|x)}\Big[ \log \tfrac{p(y|x)}{q_1(y|x)}+\log \tfrac{p(y|x)}{q_2(y|x)}\Big]\\  
		&= 2\E_{y\sim p(\cdot|x)}\Big[ \log \tfrac{p(y|x)}{\sqrt{q_1(y|x)q_2(y|x)}}\Big]\\  
		&= 2\log (1/Z(x)) ,
	\end{align*}
	where the last step follows by the definition of $Z(x)$.
	
	The proof is then completed with the following bound on the partition
	function $Z(x)$ of Algorithm~\ref{alg:CP}:
	
	\begin{lemmarep}\label{lem:partitionfunc}
		We have that:
		\[
		Z(x) = \begin{cases}
			1 -\TV(q_1(\cdot|x),q_2(\cdot |x))   & \textrm{if } \Delta=\divmax  \\
			1 -\Hel^2(q_1(\cdot|x),q_2(\cdot |x)) & \textrm{if } \Delta=\divkl.
		\end{cases}
		\]
	\end{lemmarep}
	\begin{toappendix}
		\label{app:partitionfunc}
	\end{toappendix}
	\begin{proof} 
		In Algorithm~\ref{alg:CP} with $\Delta=\divmax$ we have $p(y|x) = \frac{m(y)}{Z(x)}$ where $m(y) = \min\{q_1(y|x), q_2(y|x)\}$. Hence, $Z(x) = \sum_y m(y)$. For every $y$, $|q_1(y|x)-q_2(y|x)| = (q_1(y|x)-m(y)) + (q_2(y|x)-m(y))$, since depending on whether $q_1(y|x)>q_2(y|x)$ or vice versa, one of the terms is $|q_1(y|x)-q_2(y|x)|$ and the other is zero. 
		Hence,
		\[
		2\TV(q_1(\cdot|x),q_2(\cdot|x)) = \sum_y |q_1(y|x)-q_2(y|x)| = \sum_y q_1(y|x)+q_2(y|x) -2m(y) = 2 - 2\sum_{y} m(y)
		\]
		
		which implies that
		
		$$Z(x) = \sum_y m(y) = 1 -\TV(q_1(\cdot|x),q_2(\cdot |x))$$
		
		which is what we needed to prove.

		In Algorithm~\ref{alg:CP} with $\Delta=\divkl$ we have $p(y|x) = \frac{\sqrt{q_1(y|x)q_2(y|x)}}{Z(z)}$.  So
		$$Z(x) = \sum_y \sqrt{q_1(y|x)q_2(y|x)} = 1-\Hel^2(q_1(\cdot|x),q_2(\cdot|x))$$
		
		where the last equality follows from the definition of $\Hel^2$.
	\end{proof}
	
	For $\Delta=\divmax$, the proof of this lemma follows from a standard probability mass
	argument using properties of the total variation distance, and, for the
	$\divkl$ case, the lemma directly follows from the definition of the
	Hellinger distance. See the Appendix~\ref{app:partitionfunc}.
	
	\paragraph{Bounded degradation.}
	
	Our goal is to not only prevent copyright infringement but, importantly, to also maintain high-quality generative models when $\cA(\cD)$ itself is a high-quality model. 
	The following lemma formalizes this, showing that \CPdelta does not substantially degrade the quality of the model (in comparison to a model trained on half the data).
	
	\begin{lemmarep}[Bounded Degradation] \label{lem:boundeddegradation}
		Let $p$ be the model returned by \CPdelta, and $q_1$ and $q_2$ be the models returned by \ssafe.
		For $i\in \{1,2\}$ and for $\Delta=\divmax$,
		\[
		\TV\bigl( p(\cdot|x)  ,  q_i(\cdot|x) \bigr) \leq
		\TV\bigl( q_1(\cdot|x)  ,  q_2(\cdot|x) \bigr),
		\]
		and for $i\in \{1,2\}$ and for $\Delta=\divkl$,
		\[
		\KL\bigl( p(\cdot|x)  ,  q_i(\cdot|x) \bigr) \leq
		- 2\cdot \log \left( 1-\Hel^2\bigl( q_1(\cdot|x)  ,  q_2(\cdot|x) \bigr)  \right).
		\]
	\end{lemmarep}
	\begin{proof}
		For $\Delta=\divmax$ by Lemma~\ref{lem:partitionfunc} we have that $p(y|x) = \frac{\min\{q_1(y|x), q_2(y|x)\}}{1-\TV\bigl( q_1(\cdot|x)  ,  q_2(\cdot|x) \bigr)}$ . Hence,
		
		\begin{align*}
			\TV\bigl( p(\cdot|x)  ,  q_1(\cdot|x) \bigr) &= \frac{1}{2}\sum_y \abs*{q_1(y|x) - \frac{\min\{q_1(y|x), q_2(y|x)\}}{1-\TV\bigl( q_1(\cdot|x)  ,  q_2(\cdot|x) \bigr)}} \\
			&= \frac{1}{2}\sum_y \abs*{q_1(y|x) - \min\{q_1(y|x), q_2(y|x)\}-\frac{\TV\bigl( q_1(\cdot|x)  ,  q_2(\cdot|x) \bigr)\min\{q_1(y|x), q_2(y|x)\}}{1-\TV\bigl( q_1(\cdot|x)  ,  q_2(\cdot|x) \bigr)}} \\
			&\leq \frac{1}{2}\sum_y\max\left\{q_1(y|x) - \min\{q_1(y|x), q_2(y|x)\}, \frac{\TV\bigl( q_1(\cdot|x)  ,  q_2(\cdot|x) \bigr)\min\{q_1(y|x), q_2(y|x)\}}{1-\TV\bigl( q_1(\cdot|x)  ,  q_2(\cdot|x) \bigr)}\right\} \\
			&\leq \frac{1}{2}\left(\sum_y q_1(y|x) - \min\{q_1(y|x), q_2(y|x)\} + \sum_y\frac{\TV\bigl( q_1(\cdot|x)  ,  q_2(\cdot|x) \bigr)\min\{q_1(y|x), q_2(y|x)\}}{1-\TV\bigl( q_1(\cdot|x)  ,  q_2(\cdot|x) \bigr)}\right)
		\end{align*}
		
		where the second last inequality follows from the fact that for $a \geq 0, b \geq 0$ we have $|a-b| \leq \max\{a, b\}$. 
		
		Using arguments similar to those used in the proof of Lemma~\ref{lem:partitionfunc} it is easy to see that 
		\[ \sum_y q_1(y|x) - \min\{q_1(y|x), q_2(y|x)\} = \TV(q_1(y|x), q_2(y|x)) \] 
		
		and 
		\[ \sum_y\frac{\TV\bigl( q_1(\cdot|x)  ,  q_2(\cdot|x) \bigr)\min\{q_1(y|x), q_2(y|x)\}}{1-\TV\bigl( q_1(\cdot|x)  ,  q_2(\cdot|x) \bigr)} = \TV(q_1(y|x), q_2(y|x))\; . \] 
		
		Therefore, 
		
		$$\TV\bigl( p(\cdot|x)  ,  q_1(\cdot|x) \bigr) \leq \frac{1}{2}\left(\TV(q_1(y|x), q_2(y|x))+\TV(q_1(y|x), q_2(y|x))\right) = \TV(q_1(y|x), q_2(y|x))$$
		
		A symmetric statement holds for $\TV\bigl( p(\cdot|x)  ,  q_2(\cdot|x) \bigr)$ implying  that for
		$i\in \{1,2\}$ and for $\Delta=\divmax$,
		\[
		\TV\bigl( p(\cdot|x)  ,  q_i(\cdot|x) \bigr) \leq \TV(q_1(y|x), q_2(y|x)),
		\]
		
		For $\Delta=\divkl$ we have that $k_x = \max_{i \in \{1, 2\}} \KL(p(\cdot|x), q_i(\cdot|x))$ and by Theorem~\ref{thm:cpalg} we have that $k_x \leq -2 \log \left(1 - \Hel^2\bigl( q_1(\cdot|x) \,,\,q_2(\cdot|x) \bigr)\,\right)$. Hence,
		\[
		\divkl\bigl( p(\cdot|x) \,  \| \, q_i(\cdot|x) \bigr) \leq -2 \log \left(1 - \Hel^2\bigl( q_1(\cdot|x) \,,\,q_2(\cdot|x) \bigr)\,\right)
		\]
		which is what we needed to prove.

	\end{proof}
	In particular, if $q_1$ and $q_2$ are $\eps$ close to each other in total variation then $p$ will be $\eps$ close to both $q_1$ and $q_2$ in total variation. 
	Thus $p$ is not much worse in quality in comparison to $q_1$ and $q_2$. 
	(Our experiments actually show $p$ can even be higher quality, possibly due to \CPdelta having a model averaging effect). 
	The benefit now is that $p$ itself will (essentially) no longer sample copyrighted material (i.e. it does so only $1/(1-\eps) \approx 1+\eps$ times more than the safe model).
	As an illustrative case, suppose $q_1,q_2$ are $0.01$ close to each other in total variation, but this $1\%$ difference corresponds to a chance that each model $q_i$ outputs, verbatim, some copyrighted material that is not accessed (in training) by the other model.
	Hence, we expect the probability under $q_2$ of outputting copyrighted material output by $q_1$ to be exponentially small, and vice versa.
	Our algorithm transforms these models into a new model $p$, which is $\approx .01$ close to either of the original models in total variation, but $p$ now outputs copyrighted material with an exponentially small probability. 
	While a $1\%$ performance degradation may be tolerable in many settings, outputting copyrighted material $1\%$ of the time is very likely not acceptable (e.g. due to resulting liabilities).

	\subsection{Black-Box Oracle Algorithms}\label{sec:cpk}
	
	There are a number of modifications worth considering for practical deployments.
	First, implementing \CPdelta may be
	computationally difficult in practice, e.g. if $q_1$ and $q_2$ are
	neural models for text sequences or image generation.  Second, as
	\CPdelta is parameter free, it may be worthwhile to introduce a
	parameter based version for greater protection. Finally, it may be
	desirable to use a reduction based approach, where we can more directly
	modify any model $p$ to make it access-free, e.g. we may want to 
	directly utilize a model $p=\cA(\cD)$ that was trained on all of the data.

	Let us say $\cV=\{q_1, \ldots q_n\}$ is a cover of the function \safe
	if for all $C\in\cC$, there exists some $q\in \cV$ such that
	$\safe(C)=q$. 
	Section~\ref{sec:largerm} provides a construction for a \safe
	function leading to a cover whose size is greater than $2$, for the
	$m>1$ case.   The \CPk Algorithm, presented in 
	Algorithm~\ref{alg:CPk}, takes as input any model $p$, a cover $\cV$,
	and a threshold $k$ and returns a model $p_k$, which has quantifiable
	access-free guarantees with respect to \safe. \CPk assumes access
	to an oracle where we can both compute conditional probabilities and
	obtain samples under the models
	$p$ and $q\in\cV$.
	
	\begin{algorithm}
		\centering
		\caption{\CPk} \label{alg:CPk}
		\begin{minipage}{5in}
			\begin{algorithmic}  
				\BeginBox
				\Procedure{\CPk: \textnormal{Access-Free Reduction at Threshold $k$}}{} 
				\State \textbf{Input:} a model $p$; a cover $\cV$ of \safe;
				a threshold $k\geq 0$.
				\State \textbf{Return $p_k$:} where $p_k$ is specified as:
				\State \While{True}: Sample $y \sim
				p(\cdot|x)$ and accept $y$ if,
				\begin{equation} \label{eq:kthreshold}
					\forall q\in \cV, \  \log \big( p(y|x)/q(y|x) \big) \leq k.
				\end{equation} 
				\State \EndWhile
				\EndProcedure
				\EndBox
			\end{algorithmic}
		\end{minipage}
	\end{algorithm}

	The intuition of \CPk is as follows: we first sample
	$y\sim p(\cdot|x)$ and only accept this output if it satisfies a
	desired upper bound with regards to the function \safe, the latter of
	which can be efficiently checked using the cover $\cV$ of \safe. One
	potentially undesirable property of this algorithm is that it is
	discontinuous: an output with probability slightly above the
	acceptance threshold in (\ref{eq:kthreshold}) will be rejected.  The
	\CPsmooth Algorithm, presented in Algorithm~\ref{alg:CPsmooth},
	provides a modification where the acceptance probability is a
	continuous function of $p(\cdot|x)$ (leading to a slightly improved
	efficiency bound in Theorem~\ref{thm:cpk_efficiency}).

	Let $\nu_k(x)$ be the probability that $y$ is accepted on input $x$
	in any iteration of the while statement in \CPk or \CPsmooth (the attempts are i.i.d.). The
	guarantee for both algorithms are as follows:
	
	\begin{theoremrep}[Guarantees for \CPk and \CPsmooth] \label{thm:cpk}
		Let $p_k$ be the model returned by either \CPk or \CPsmooth when input with a model
		$p$; a cover $\cV$ for \safe, and a threshold $k\geq 0$.
		Let $\nu_k(x)$ be the probability that the sampled $y$ is accepted in a single iteration of the while loop.
		We have that:
		\begin{itemize}
			\item (Near Access-Freeness) $p_k$ is $\widetilde k_x$-NAF on prompt $x$ with
			respect to $\safe$ and $\Delta=\divmax$, where:
			\[
			\widetilde k_x \leq k+\log(1/\nu_k(x)).
			\]
			\item (Model Degradation) $p_k$ satisfies the following bound:
			\[
			\TV\bigl( p_k(\cdot|x)  ,  p(\cdot|x) \bigr)  \leq 1-\nu_k(x).
			\]  
			\item (Oracle Complexity) Sampling $y\sim p_k(\cdot|x)$ requires $O(1/\nu_k(x))$ iterations, where each iteration involves obtaining one sample from $p$ and doing $|\cV|+1$ probability computations from either $p$ or $q\in\cV$.
		\end{itemize}
	\end{theoremrep}
	\begin{proof}
		We start with bounding $\widetilde k_x$. For \CPk $y$ is sampled in a single iteration of the while loop if $p(y|x) \leq \min_{q \in \cV} 2^k q(y|x))$. Hence, the probability of sampling $y$ in a single iteration of while statement is $\leq \min_{q \in \cV} 2^k q(y|x))$.
		For \CPsmooth the probability of sampling $y$ in a single iteration of while statement is $\leq \min\{p(y|x), \min_{q \in \cV} 2^k q(y|x))\} \leq \min_{q \in \cV} 2^k q(y|x))$. Hence, for both \CPk and \CPsmooth the probability of sampling $y$ in a single iteration of while statement is $\leq \min_{q \in \cV} 2^k q(y|x))$.  
		
		This implies that the overall probability of sampling $y$ i.e $p_k(y|x)$ is $ \leq \min_{q \in \cV} 2^k q(y|x))/\nu_k(x)$. By definition of NAF we have that $p_k$ is $\widetilde k$-NAF where $\widetilde k = \max_{y, q \in \cV} \log(p(y|x)/q(y|x))$. Using $p_k(y|x) \leq \min_{q \in \cV} 2^k q(y|x))/\nu_k(x)$ we get that $\widetilde k \leq \log(2^k/\nu_k(x)) = k + \log(1/\nu_k(x))$.
		
		We now move to bounding model degradation. For \CPk, since $p_k(\cdot|x)$ is just renormalized $p(\cdot|x)$ on a subset with mass $\nu_k(x)$ we have that
		\begin{align*}
			\TV\bigl( p_k(\cdot|x)  ,  p(\cdot|x) \bigr) &= \sum_{y, p_k(y|x) > p(y|x)} (p_k(y|x) - p(y|x))\\
			&= \sum_{y, p_k(y|x) > p(y|x)} (p(y|x)/\nu_k(x) - p(y|x))\\
			&= \sum_{y, p_k(y|x) > p(y|x)} p(y|x)(1/\nu_k(x) - 1)\\
			&= (1/\nu_k(x) - 1)\sum_{y, p_k(y|x) > p(y|x)} p(y|x)\\
			&=\left(\frac{1}{\nu_k(x)}-1\right)\nu_k(x)\\
			&= 1-\nu_k(x).
		\end{align*}
		
		For \CPsmooth, let $w_x(y) = \min\{p(y|x), 2^k\min_i(q_i(y|x))\}$. Note that $p_k(y|x) = w_x(y)/\nu_k(x)$ and $\sum_y w_x(y) = \nu_k(x)$. We have that,
		\begin{align*}
			\TV\bigl( p_k(\cdot|x)  ,  p(\cdot|x) \bigr) &= \sum_{y, p_k(y|x) > p(y|x)} (p_k(y|x)-p(y|x))\\ 
			&\leq \sum_{y, p_k(y|x) > p(y|x)} (p_k(y|x)-\min\{p(y|x), 2^k\min_i(q_i(y|x))\})\\
			&= \sum_{y, p_k(y|x) > p(y|x)} (p_k(y|x)-w_x(y))\\
			&= \sum_{y, p_k(y|x) > p(y|x)} (w_x(y)/\nu_k(x)-w_x(y))\\
			&\leq \sum_{y} (w_x(y)/\nu_k(x)-w_x(y))\\
			&= \left(\frac{1}{\nu_k(x)}-1\right)\sum_{y} w_x(y)\\
			& = \left(\frac{1}{\nu_k(x)}-1\right)\nu_k(x) = 1-\nu_k(x).
		\end{align*}
	\end{proof}
	
	Because $k$ appears on the right-hand side of (\ref{eq:kthreshold}), the higher we set the parameter $k$, the higher the probability  $\nu_k(x)$ that $y$ is accepted.
	Hence, making $\widetilde k_x$ acceptably small involves balancing the two components.
	One heuristic would be to choose $k$ as the \emph{median}  of the left-hand side of (\ref{eq:kthreshold}), which would ensure that $\nu_k(x)=1/2$, hence loosing only an additive factor of $1$ in the bound on $\widetilde{k}_x$; here, $p_k$ could then substantially provide different samples in comparison to $p$. Using a percentile instead of the median is a natural way to tune this tradeoff.
	
	As before, by Lemma~\ref{lem:nocopying}, for the case of $\divmax$,  we have that
	for all $C\in\cC$ and events $\cE$, 
	\[
	p_k(\cE|x) \, \leq \, \frac{2^k}{\nu_k(x)} \cdot \safe_C(\cE|x),
	\]
	which also shows the tradeoff in $k$.
	
	Now let us understand the efficiency of this approach and also
	consider a few natural choices for $p$, e.g. choosing $p=q_1$ itself or
	choosing $p=\cA(\cD)$, the model trained with all the data.
	
	\begin{algorithm}
		\centering
		\caption{\CPsmooth} \label{alg:CPsmooth}
		\begin{minipage}{4.3in}
			\begin{algorithmic}
				\BeginBox
				\Procedure{\CPsmooth: \textnormal{Smoothed  access-free reduction}}{} 
				\State \textbf{Input:} a model $p$; a cover $\cV$ of \safe; a threshold
				$k\geq 0$ .
				\State \textbf{Return $p_k$:} where $p_k$ is specified as:
				\State \While{True}: Sample $y \sim
				p(\cdot|x)$ and \\
				$$\textrm{return } y \textrm{ with probability } \min\left\{1,
				\min_{q\in \cV}\left\{ \frac{2^k
					q(y|x)}{p(y|x)}\right\}\right\}.$$
				\State \EndWhile
				\EndProcedure
				\EndBox
			\end{algorithmic}
		\end{minipage}
	\end{algorithm}

	\paragraph{Efficiency.}
	As shown in Theorem~\ref{thm:cpk}, the quantity $\nu_k(x)$ is critical because it governs $\widetilde k_x$, the model degradation, and the oracle complexity. We now characterize $\nu_k(x)$ based on
	a particular ``distance'' measure between $p$ and the set $\cV$. 
	In the extreme case, where
	$p(\cdot|x)$ and all $q(\cdot|x)\in\cV$ are equal to each
	other, then $p_k=p$ and the sampling succeeds at every attempt,
	i.e. $\nu_k(x)=1$. Let us now quantify the impact of when these distributions
	are not all equal to each other. Define:
	\[
	d_x(p,\cV) =  \sum_{y\in\cY}\big| p(y|x)  -  \min\{q_1 (y|x), \ldots q_n(y|x)\} \big|_+ \,
	\]
	where $|\cdot|_{+}$ is the function which thresholds negative inputs to
	$0$. 
	It is straightforward to observe that $0\leq d_x(p,\cV) \leq
	1$. For an interpretable upper bound on $d_x(p,\cV)$, we have that:
	\[
	d_x(p,\cV) 
	\leq \sum_{q\in\cV} \TV\bigl( p(\cdot|x)  ,  q(\cdot|x) \bigr),
	\]
	which shows that $d_x(p,\cV)$ will be small if $p$ and all $q\in\cV$ are close to each other.
	
	The following theorem presents our characterization of the efficiency of
	\CPk and \CPsmooth, through bounding $\nu_k(x)$.  
	
	\begin{theoremrep}[Bounds on $\nu_k(x)$] \label{thm:cpk_efficiency}
		Fix a model $p$, a function \safe, and a prompt $x$. Let $\cV=\{q_1,\ldots q_n\}$ be a cover
		for \safe. Let $d=d_x(p,\cV)$ and  assume  $d <1$. Let $p_k$ be the model returned by either \CPk or \CPsmooth with input $p$, $\cV$, and a threshold $k$. We have that:
		\begin{itemize}
			\item For \CPk and provided $k \geq  \log\big(2/(1-d)\big)$, the acceptance probability is bounded as:
			\[
			\nu_k(x) \geq \frac{1-  d}{1+d}.
			\]  
			\item For \CPsmooth and for $k \geq 0$, the acceptance probability is bounded as:
			\[
			\nu_k(x) \geq 1-  d.
			\]
		\end{itemize}
	\end{theoremrep}
	\begin{proof}
		Let $\cE_x$ be the event that
		a sample $y$ is rejected and $m_x(y) = \min\{q_1 (y|x), \ldots q_n(y|x)\}$. For \CPk, as sample $y$ is rejected i.e. $y \in \cE_x$ if and only if:
		\[
		p(y|x) \geq 2^k  \min\{q_1 (y|x), \ldots q_n(y|x)\} = 2^km_x(y)
		\]
		and so, summing over $y\in \cE_x$, leads to:
		\begin{align*}
			p(\cE_x|x) &\geq 2^k \sum_{y \in \cE_x} m_x(y)\\
			&= 2^k \sum_{p(y|x) \geq 2^km_x(y)} m_x(y)\\
			&= 2^k \sum_{p(y|x) \geq 2^km_x(y)} p(y|x)-(p(y|x)-m_x(y))\\
			&= 2^k (p(\cE_x)-\sum_{p(y|x) \geq 2^km_x(y)} (p(y|x)-m_x(y)))\\
			&\leq 2^k (p(\cE_x)-|p(y|x)-m_x(y)|_{+})\\
			&= 2^k (p(\cE_x)  - d)
		\end{align*}
		
		Rearranging, we have:
		\[
		p(\cE_x)\leq \frac{2^k}{2^k-1} d .
		\]
		Therefore,
		\[
		\nu_k(x)  =1- p(\cE_x)\geq 1-\frac{2^k}{2^k-1} d ,
		\]
		and our setting of $k =  \log\big(2/(1-d)\big)$ gives:
		\[
		1-\frac{2^k}{2^k-1} d = 1- \frac{2}{1+d} d
		= \frac{1-d}{1+d}.
		\]
		which is our claimed bound on $\nu_k(x)$.

		For \CPsmooth a sample $y$ is rejected only if $p(y|x) > 2^km_x(y)$ and in that case it is rejected with probability $p(y|x)-p(y|x)\cdot\frac{2^km_x(y)}{p(y|x)} = p(y|x)-2^k m_x(y)$. Hence we have:
		\begin{equation}\label{eq:rhok_equality}
			p(\cE_x) = \sum_{y, p(y|x) > 2^km_x(y)}\big(p(y|x)-2^k m_x(y)\big).
		\end{equation}
		Using that $k\geq 0$,
		\[
		p(\cE_x) 
		\leq \sum_{y, p(y|x) > 2^km_x(y)}\big(p(y|x)- m_x(y)\big) 
		\leq \sum_{y}\big|p(y|x)- m_x(y|x)\big|_+ 
		= d .
		\]
		Therefore,
		\[
		\nu_k(x)  =1- p(\cE_x)\geq 1-d ,
		\]
		
		and using $\widetilde k_x = k + \log(1/\nu_k(x))$ leads to the claimed bound on $\widetilde k_x$.
	\end{proof}

	A few points are in order. 
	First, to better understand the restriction $d<1$, it is not difficult to construct a case  where $d=1$ and where the acceptance probability $\nu_k(x)$ is $0$. For example, consider a case where $|\cV|=2$ and, for all $y\in \cY$, $\min\{q_1(y|x),q_2(y|x)\}=0$. Furthermore, in such a case, there exists no distribution $p$ which is $k$-NAF (for finite $k$) with respect to this \safe function. 
	Second, while the above bound on $\nu_k(x)$ is not shown to be
	increasing in $k$, we expect this to happen in practice (see
	(\ref{eq:rhok_equality}) in the Appendix~\ref{app:cpk_proofs} for a sharp expression for $\nu_k(x)$, which does increase with $k$).

	Let us consider the special case where our cover has
	two elements, i.e. $\cV=\{q_1,q_2\}$, as would be the case if we used
	\ssafe, and where we choose $p=q_1$. In such a case, the following corollary shows that
	\CPsmooth is as effective as
	\CPdelta, for $\Delta=\divmax$ (and \CPk looses a constant additive factor in $\widetilde k_x$).
	
	\begin{corollaryrep} Suppose $\cV=\{q_1,q_2\}$ is a cover
		of \safe (e.g. if \ssafe is used). Let $p=q_1$. We have:
		\[
		d_x(p,\cV) = \TV\bigl( q_1(\cdot|x)  ,  q_2(\cdot|x) \bigr).
		\]
		Therefore, the claims in Theorem~\ref{thm:cpk_efficiency} hold with
		$d=\TV\bigl( q_1(\cdot|x)  ,  q_2(\cdot|x) \bigr)$. 
		This implies that, for $k = 0$ and for \CPsmooth, we recover the guarantees of $\CPdelta$ with $\Delta = \divmax$. Furthermore, in this case, we have that $p_k$ is equal to the distribution $\min\{q_1(y|x),
		q_2(y|x)\}/Z(x)$ itself.
	\end{corollaryrep}
	\begin{proof}
		The first claim follows from
		\begin{align*}
			d_x(p,\cV) &= \sum_{y}|p(y|x) - \min\{q_1(y|x), q_2(y|x)\}|_{+}\\
			&= \sum_{y}|q_1(y|x) - \min\{q_1(y|x), q_2(y|x)\}|_{+}\\
			&= \sum_{y}(q_1(y|x) - \min\{q_1(y|x), q_2(y|x)\})\\
			&= \sum_{y, q_1(y|x) > q_2(y|x)}(q_1(y|x) - q_2(y|x))\\
			&= \TV\bigl( q_1(\cdot|x)  ,  q_2(\cdot|x) \bigr)
		\end{align*}
		
		Substituting $k = 0$ and $d=\TV\bigl( q_1(\cdot|x)  ,  q_2(\cdot|x) \bigr)$ in the guarantees of \CPsmooth from Theorem~\ref{thm:cpk},~\ref{thm:cpk_efficiency} we get that,
		
		\begin{align*}
			\widetilde k_x &\leq 0+\log(1/(1-\nu_k(x)))\leq\log(1/(1-d))=-\log(1- \TV\bigl( q_1(\cdot|x)  ,  q_2(\cdot|x) \bigr))\\
			\TV\bigl( p_k(\cdot|x)  ,  p(\cdot|x) \bigr) &\leq 1-\nu_k(x) \leq 1-(1-d)=d=\TV\bigl( q_1(\cdot|x)  ,  q_2(\cdot|x) \bigr)
		\end{align*}
		
		which are exactly the guarantees we obtained from \CPdelta for $\Delta = \divmax$. This is not a coincidence since we now show that in this case $p_k(y|x) = \min\{q_1(y|x),
		q_2(y|x)\}/Z(x)$. For $k = 0$ the probability of sampling $y$ in a single iteration of while loop in \CPsmooth is $$\min\{p(y|x), 2^k\min_{i \in \{1, 2\}}\{q_i(y|x)\}\} = \min\{q_1(y|x), 2^0\min_{i \in \{1, 2\}}\{q_i(y|x)\}\} = \min\{q_1(y|x),
		q_2(y|x)\}.$$ Normalizing this gives us that $p_k(y|x) = \min\{q_1(y|x),
		q_2(y|x)\}/Z(x)$.

	\end{proof}
	
	Provided $d$ is non-trivially bounded away from $1$, say $d=1-\delta$,
	we expect this to be a strong guarantee on the violation probability (as per the discussion in
	Section~\ref{sec:knaf}), though now $\nu_k(x)$ may be as small as $\delta$, making the
	sampling costly for very small $\delta$. However,
	for moderately small values of $\delta$, both algorithms will be
	efficient to implement.
	
	\begin{toappendix}
		\label{app:cpk_proofs}
		
	\end{toappendix}
	
	\subsection{Handling Multiple Accessing Datapoints: The $m>1$ Case.} \label{sec:largerm}
	
	\begin{algorithm}
		\caption{\ssafe, $m>1$ case} \label{alg:ssafe_general}
		\centering
		\begin{minipage}{4.6in}
			\begin{algorithmic} 
				\BeginBox
				\Procedure{Sharded Safe \textnormal{($m>1$ case)}}{} 
				\State \textbf{Input:}  Dataset $\cD$
				\State \textbf{Shard $\cD$:} Partition $\cD$ into
				$\cD_1, \ldots \cD_{m+1}$ datasets (see text).
				\State \textbf{Learning:} For $i=1, \ldots m+1$, set $q_i =
				\cA(\cD_i)$.
				\State \textbf{Specification:} For all $C \in
				\cC$, \\
				Set $\ssafe(C) = q_i$, for $i$ s.t. $C\notin \cD_i$.
				\State \textbf{Return:}  Models $q_1, \ldots
				q_{m+1}$ and $\ssafe$.
				\EndProcedure
				\EndBox
			\end{algorithmic}
		\end{minipage}
	\end{algorithm}

	Recall that $m$ denotes the number of datapoints that can access a single copyrighted data $C$.
	When $m>1$, it may be the case that a datapoint $z\in\cD$ has copyrighted more
	than one work in $\cC$ (e.g. some training datapoint substantially
	contains material from two copyrighted works), so simply deduplicating $\cD$ may not result in a
	dataset with $m=1$. Hence, an algorithm for $m>1$ may be desired.
	The more general $\ssafe$ Algorithm is presented in
	Algorithm~\ref{alg:ssafe_general}. 
	The algorithm $\ssafe$ first partitions $\cD$ into disjoint shards $\cD_1, \ldots \cD_{m+1}$.
	By the pigeonhole principle, this ensures that each copyrighted work does not appear in at least one dataset $\cD_i$. Of course, depending on the dataset, it may be possible to use less than $m+1$ partitions, even for the $m>1$ case.
	
	
	The guarantees for \CPk Algorithm already extend to using 
	\ssafe; for example by taking $p=\cA(\cD)$, we then have a black box
	procedure for copyright protection using only the algorithm
	$\cA$.
	Even though \CPk is a natural algorithm to use, it may still be
	conceptually worthwhile to modify the \CPdelta algorithm to handle the
	$m>1$ case Here,
	we see show how a certain log partition
	function governs $k_x$. For $\Delta=\divmax$, \CPdelta can be modified to return:
	\[
	p(y|x) = \frac{\min\{q_1(y|x), \ldots, q_{m+1}(y|x)\}}{Z(x)},
	\]
	and, for $\Delta=\divkl$, \CPdelta can be modified to return:
	\[
	p(y|x) = \frac{\big(  q_1(y|x) q_2(y|x) \ldots q_{m+1}(y|x) \big)^{1/(m+1)}}{Z(x)}.
	\]
	
	This algorithm enjoys the following guarantee:
	
	\begin{lemmarep}[\CPdelta, $m>1$] \label{lemma:maxklalg_general}
		Let $p$ be the model defined above.
		We have that $p$ is $\widetilde k_x$-NAF with respect to $\ssafe$, where
		$ \widetilde k_x \leq -  \log Z(x) $
		if $\Delta = \divmax$ and $\widetilde k_x \leq -  (m+1)\log Z(x)$ if $\Delta = \divkl$.
	\end{lemmarep}
	\begin{proof}
		First, for $\Delta=\divmax$, observe that, by construction, $p(y|x) \leq q_i(y|x)/Z(x)$ 
		for all $y\in\cY$, $i\in\{1,\dots,m+1\}$. Hence, $k_x = \max_{i, y} \log(p(y|x)/q_i(y|x))
		\leq \log(1/Z(x)) = -\log Z(x)$. 
		
		For $\Delta=\divkl$, we have that 
		\begin{align*}
			k_x &= \max_{i \in \{1, \ldots,m+1\}} \KL(p(\cdot|x), q_i(\cdot|x))\\
			&\leq \sum_{i=1}^{m+1} \KL(p(\cdot|x), q_i(\cdot|x))\\
			&= (m+1)\E_{i\in\{1,\ldots,m+1\}} \KL(p(\cdot|x), q_i(\cdot|x))\\
			&= (m+1)\E_{y\sim p(\cdot|x)} \log \frac{p(y|x)}{(q_1(y|x)\cdot q_2(y|x)\ldots q_{m+1}(y|x))^{1/(m+1)}}\\
			&= (m+1)\E_{y\sim p(\cdot|x)} \log \frac{1}{Z(x)}\\
			&= -(m+1)\log Z(x)
		\end{align*}
		\vspace*{-0.1cm} 
	\end{proof}
	The proof is analogous to the $m=1$ case and is provided in the Appendix~\ref{app:cpk_proofs}.

	\section{Experiments} \label{sec:empirical}

	We now provide experimental validation for both language and image generative models.
	While there is significant room for the optimization of this approach and for the use of large datasets, this is not our focus. Instead, our experiments are for validation and demonstrating that our algorithms lead to minimal performance degradation while providing rigorous bounds on the distance from the access-free models.
	Qualitatively, we also observe that applying our algorithm can transform models, each of which has significant chance of outputting some fixed memorized element, into a combined model where this probability is greatly reduced.
	These experiments should be considered as proof of concept, meant to highlight that the approach is both viable and simple to implement. There are several natural modifications for reducing the quantitative bounds on $k_x$ as well as improving performance, which we leave to future work. 
	All our experiments use the \ssafe function (Algorithm~\ref{alg:ssafe}).
	That is, we split the dataset $\cD$ into two disjoint parts $\cD_1$ and $\cD_2$, and train two separate models $q_1,q_2$ on those.

	\begin{figure*}[t]
		\centering
		\minipage{0.45\textwidth}
		\includegraphics[width=\linewidth]{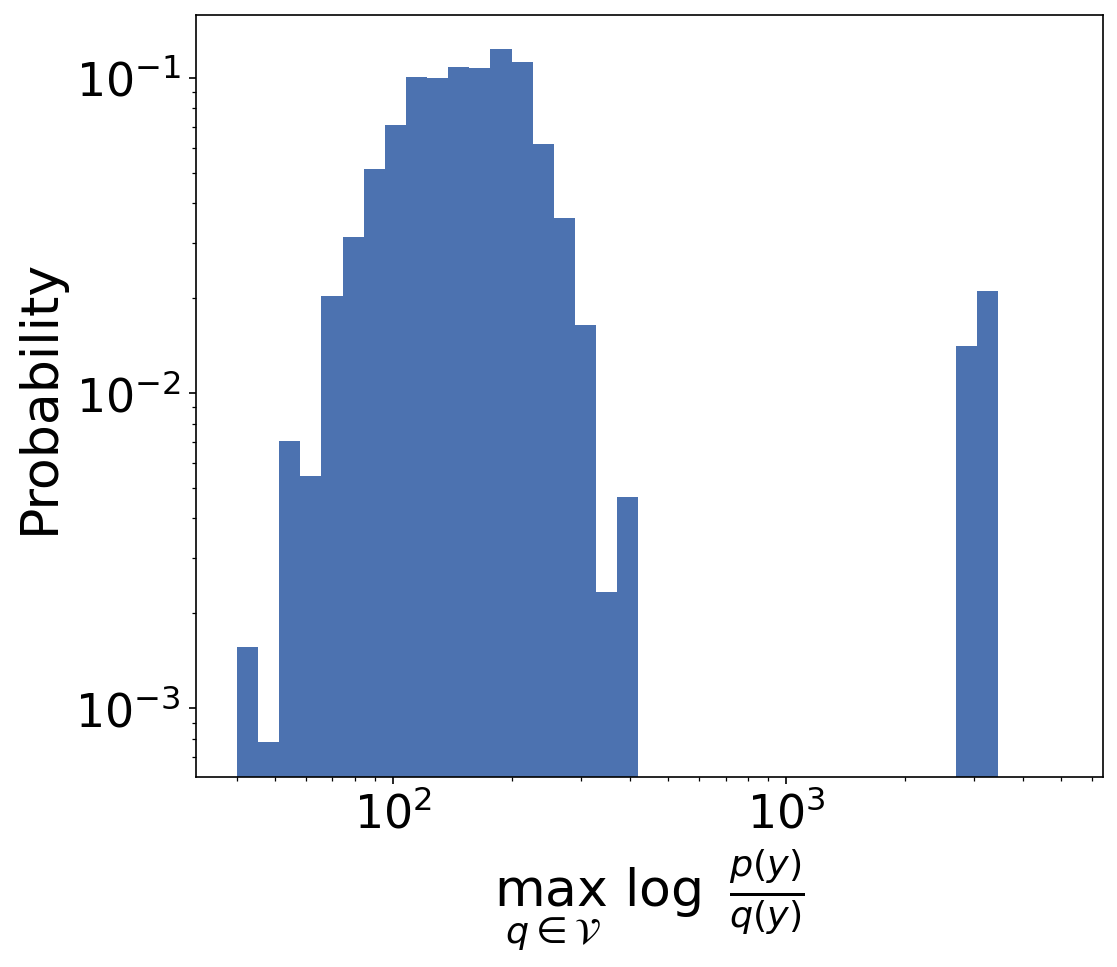}
		\endminipage\hfill
		\minipage{0.45\textwidth}%
		\includegraphics[width=\linewidth]{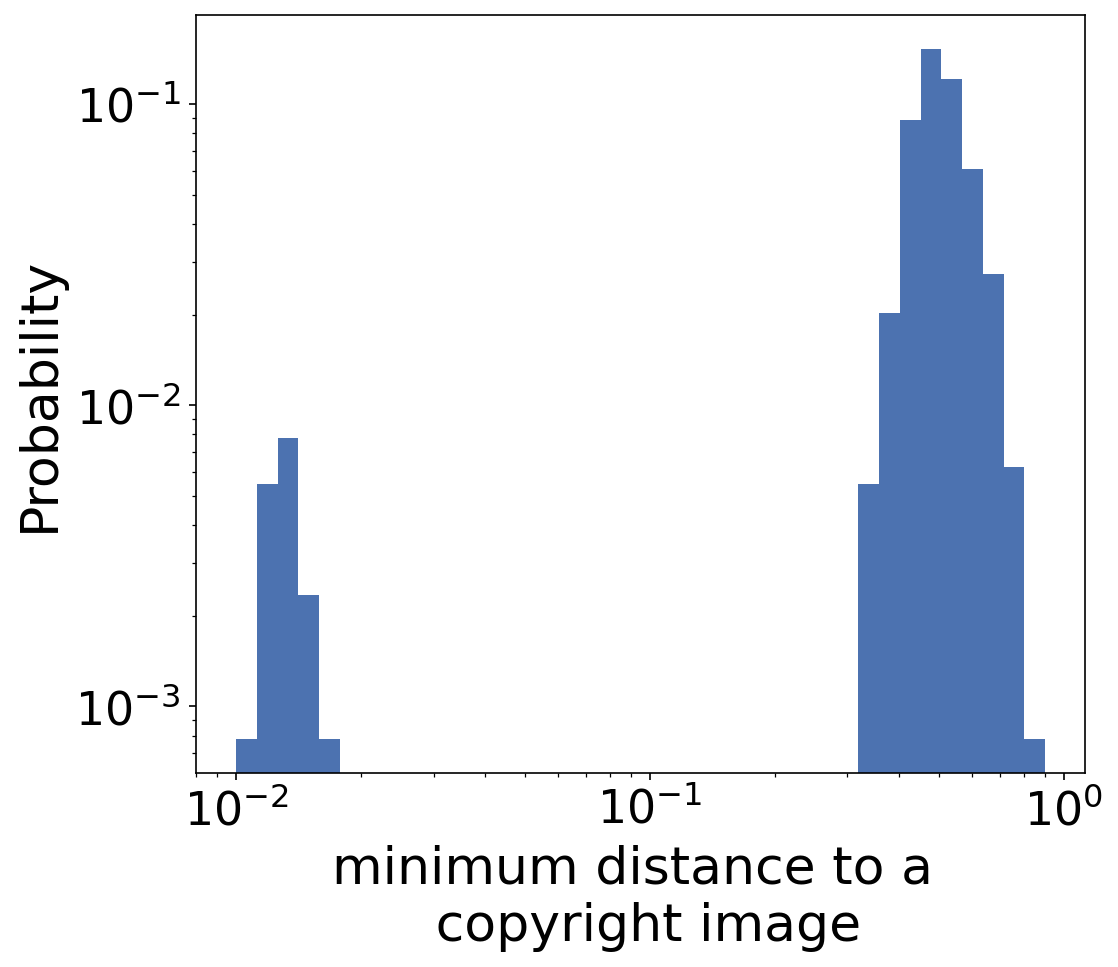}
		\endminipage
		\caption{Histograms associated with the diffusion experiment. }
		\label{fig:diffusion-histograms}
	\end{figure*}

	\subsection{A Diffusion Model Experiment} 
	We train U-net based diffusion
	models (specifically based on~\citet{cifar-diffusion}), which when trained on the full CIFAR-10 dataset (along with horizontal flips) achieves an FID score of 3.28.  The dataset we use is CIFAR-10 (along with horizontal flips) augmented with multiple copies of two images taken from the CIFAR-10 test set\footnote{The test images are visually far from each other, see Figure~\ref{fig:diffusion-intro-fig}. We also chose images which did not have duplicates in the CIFAR-10 train test, using data from~\cite{BarzD20}.}  (images close in $\ell_1$ to one of the augmented images are marked with red boundaries in Figure~\ref{fig:diffusion-intro-fig}); hypothetically, suppose these two images are copyrighted works. These two augmented images make up about $2\%$ of the new training dataset (i.e. of 50k). The leftmost image shows generations from a model $p$ that was trained on the full dataset, where we clearly see that $p$ generates the two copyrighted works. 
	Our algorithm starts by splitting this dataset into two
	disjoint datasets, making sure that copyrighted images are split into two different shards. For
	illustrative purposes, we do not deduplicate the dataset. 
	
	We now present experiments with \CPk using a threshold of $k=500$ to obtain the model $p_k$. 
	Example outputs of the four models are shown in Figure~\ref{fig:diffusion-intro-fig}.
	We see that  $p_k$ does \emph{not} output the copyrighted images with significant probability.  Even though the threshold of $k$ is large, the effect of our transformation on the probability of outputting a copyrighted image is easily noticeable. We sampled 2000 images from the transformed model and found that none of them violate copyright, 
	verifying this both visually and using the $\ell_1$ distance to the copyrighted images (discussed in the next paragraph).
	
	We now examine the effect of thresholding. Let $y = (x_T, x_{T-1}, \ldots, x_0)$ be the full reverse diffusion process for an output image $x_0$. Figure~\ref{fig:diffusion-histograms} (left) plots the histogram of $\max_{i \in \{1, 2\}}(\log(p(y)/q_i(y))$ over sampled trajectories $y\sim p$ of the reverse diffusion process. We note that this distribution is clearly bimodal. Let us denote the set of images in the second mode by $\mathcal{H}$. Visually we found that all images in $\mathcal{H}$ correspond to the copyrighted images while all images in the first mode correspond to other CIFAR-10-like images. To verify this, Figure~\ref{fig:diffusion-histograms} (right)
	plots the histogram of the $\ell_1$ distance of each image (for the same set of images) to the closest copyrighted image. Again, it is a bimodal distribution, and we find that the set of images occurring in the first mode of this figure, i.e. images which are close in $\ell_1$ to some copyrighted image, is exactly the set $\mathcal{H}$. These observations show that that a threshold of $k=500$ in the \CPk algorithm removes the copyrighted images while keeping the distribution over other CIFAR-10-like images similar to what it was before. 
	We find that $\nu_k = .965$ i.e. only $3.5\%$ of the images from the model distribution of $p$ are removed. The value of $\nu_k = .965$ also gives us that $\widetilde k = k + \log(1/\nu_k) = 500+ \log(1/.96) < 501$. 
	
	\paragraph{Training techniques to increase model similarity:} Our theoretical results (Theorem~\ref{thm:cpk_efficiency}) show that the bound on $\widetilde k$ depends on how close the underlying models $q_1, q_2$ and $p$ are. To encourage model similarity during finite-sample training, we use the same values of noise in the diffusion process (while training) for all the models (ensured by using the same random seed in training $q_1, q_2$ and $p$); this does not invalidate the access-free property of the safe models because the noise sequence is chosen independently of the training images.
	Figure~\ref{fig:diffusion-intro-fig} displays the model generations, for $p,q_1,q_2$, using the same noise sequence on the diffusion paths for the corresponding images.
	Here, we can see that these models produce similar (but not identical) images when given the same noise sequence. The rightmost figure, which shows samples from $p_k$, is exactly the same images as leftmost figure, which are samples from $p$, except when the image fails to meet the threshold criteria, in which case the image was continually re-sampled until the threshold criteria is met.

	\paragraph{The data-processing inequality and interpreting $\widetilde k$:} The value of $\widetilde k\approx 500$ may look pessimistic at first glance. A few points are in order here. First, our guarantees (Theorems~\ref{thm:cpk},\ref{thm:cpk_efficiency}) apply to the whole sequence $y$ rather than just to $x_0$, where our guarantees are on events defined on the sequences $(x_T, x_{T-1}, \ldots, x_0)$ themselves. Ultimately, we are only interested in the marginal probabilities of the images $x_0$, and, by the data-processing inequality, our bounds also hold directly on $x_0$. In particular, for any image $x_0$ generated by $p_k$, we have
	$p_k(x_0) \leq 2^{501} \cdot \safe_{x_0}(x_0)$. Part of the reason for a large value of $k$ may be due to our inability to directly run our algorithm on the marginal probabilities of the images. This is due to
	the difficulty in directly computing marginal likelihoods with diffusion models\footnote{Such an issue does not arise for language models since the whole path serves as the output i.e. there is no need for appealing to the data-processing inequality. It also would not arise for flow based (invertible) generative models}, which requires summing over different paths $y$ which end in $x_0$. 
	
	Second, it is plausible our value of $\widetilde k$ may be non-vacuous. Assuming the diffusion model is faithful to the CIFAR-10 distribution and that the negative log likelihood (NLL) of CIFAR-10 images concentrates, we have that $\safe_{x_0}(x_0) \approx 2^{-\E[\text{NLL}]}$. As the current best estimates (\cite{kingma2021on}) for $\E[\text{NLL}]$ for CIFAR-10 is around $2.5 \cdot 3072 = 7680$, we have that the probability of generating a copyrighted image, as is, is $\lesssim 2^{501} \cdot 2^{-7680} \ll 2^{-7000}$. This bound is only for generating a copyrighted image verbatim. 
	
	Finally, it is sometimes the case that theoretically grounded methods work better in practice then their bounds suggest. However, that we plausibly obtain non-vacuous bounds even when running our algorithms on the full sequence along the diffusion path is encouraging.
	
	\subsection{A Language Model Experiment}
	\begin{table}[] 
		\centering
		\begin{subtable}[h]{0.45\textwidth}
			\subcaption*{Cross entropy losses}
			\begin{tabular}{|l|l|l|l|}
				\hline
				Algorithm       & Original model & $\CPdelta_{\text{max}}$  & $\CPdelta_{\text{KL}}$\\ \hline
				350m params & 3.2      & 3.16 & 3.13            \\ \hline
				125m  params & 3.87     & 3.78 & 3.7               \\ \hline
			\end{tabular}
		\end{subtable}

		\vspace{0.5ex}

		\begin{subtable}[h]{0.45\textwidth}
			\subcaption*{Expected bounds on $k_x$}
			\begin{tabular}{|l|l|l|l|}
				\hline
				& $\E[k_x], \divmax$ &  $\E[k_x], \divkl$ & Entropy \\ \hline
				350m & .228 (7\%)              & .058 (1.7\%)              & 3.225  (100\%)  \\ \hline
				125m & .508 (13\%)              & .166   (4.5\%)            & 3.637  (100\%) \\ \hline
			\end{tabular}
		\end{subtable}
		
		\caption{Top: Cross-entropy loss for the original language generation models ($q_1$ and $q_2$) vs. ones produced by our algorithm; our algorithms improve on the loss of the underlying models. Bottom: Expectation of our bounds on $k_x$ for both $\divmax$ and $\divkl$ over the single-token distributions of text and prompts $x$ from the combined training set. (In parenthesis, the expected value of $k_x$ as a fraction of the total entropy in the token distribution.)}\label{tab:llm}
	\end{table}
	
	We use the C4 dataset~\cite{c4dataset} and train decoder-only transformers similar to GPT models (specifically~\cite{mosaic-llm}) on two disjoint parts\footnote{The amount of data used to train each $q_i$ follows the default values in~\cite{mosaic-llm}, which uses the Chinchilla (\cite{chinchilla}) compute-optimal values.} in  to obtain models $q_1,q_2$. We then transform $q_1,q_2$ to a model $p$ using the $\CPdelta$ algorithm for both $\Delta = \divmax$ and $\Delta = \divkl$. Our motivation is to understand how the $\CPdelta$ algorithm, used as is (on a token level), fares in terms of $k_x$ and the model degradation. As shown in Table~\ref{tab:llm} (top), the resulting models have somewhat improved cross-entropy loss compared to each one of the original models. For $\Delta=\divkl$, this is perhaps expected since 
	$\CPdelta$ corresponds to a model averaging algorithm in logit space.
	
	We also investigate the effectiveness of \CPdelta by looking at the implied $k_x$ at the token level. Here, we look at the expected value of $k_x$, where $x$ is a random prefix of the training data. We show that the expected value of $k_x$ is significantly smaller than the total entropy of the tokens (see Table~\ref{tab:llm} (bottom).\footnote{Theorem~\ref{thm:cpalg} uses the bound $k_x \leq  \sum_{i=1}^2 \divkl\big( p(\cdot|x) \,  \| \, q_i(\cdot|x) \big)$ in the case $\Delta=\divkl$. Instead of using that we explicitly report in Table~\ref{tab:llm} (bottom) the expectation of the true quantity $k_x = \max_{i\in \{1,2\}}(\divkl\big( p(\cdot|x) \,  \| \, q_i(\cdot|x) \big))$.}). Interestingly, even the relative bounds on the expect value of $k_x$, compared to the total entropy, improve as the model scales up, though this should be investigated for larger models.  
	
	As a crude interpretation for these per token results, assume that for all $x$, (i) $k_x$ is concentrated i.e. it is close to its expectation, and (ii)  in the safe model $q_i$, $-\log q_i(\cdot|x)$ is bounded above by some constant factor times its unconditional expectation, i.e. bounded by $O(\E_{x, y \sim q_i(\cdot|x)}[-\log q_i(y|x)])$ (which is exactly the expected entropy). Then for strings of length $\ell$ their probability in the safe model (which could be $q_1$ or $q_2$ depending on the prompt) will scale as $ \approx 2^{-\E[\text{entropy}]\cdot \ell}$, and we can bound the output probability in the final model by $ \approx 2^{\E[k_x] \ell} \cdot 2^{-\E[\text{entropy}]\cdot \ell}$. As $\E[k_x] < \E[\text{entropy}]$ this implies that the probability of violating a copyright goes down exponentially with the length of the string. While these assumptions will not strictly hold in practice, this is a starting point to understand the effectiveness of our algorithms on language models.  Further investigations can be done by directly examining the behavior of \CPk when applied to the joint probabilities on strings.

	\section{Related works}\label{sec:prior}

	There have been several studies of copyright issues in machine learning and data mining in the law literature, though most of them focus on potential infringements in the \emph{training phase}.
	\citet{sag2018new} surveys the question of whether data mining and machine learning on copyrighted text falls under ``fair use'' and states that ``allowing [text data mining] and other similar non-expressive uses of
	copyrighted works without authorization is entirely consistent with the fundamental structure of copyright law.''. \citet{sag2018new} also states that under U.S. law ``extracting a short phrase or snippet of text from
	one work and using it in another does not amount to a reproduction of the work if the localized similarity is not substantial, is not quantitatively or qualitatively significant, or is otherwise de minimis.'' (However, European courts have a stricter threshold for the amount of similarity.) \citet{sobel2018artificial} also discusses the issue of ``fair use'' in training. While he mentions the issue of output generation, the article does not focus on it since (at the time) ``works generated by Al are fascinating and entertaining, but today they remain novelties rather than mainstream sources of entertainment or compelling substitutes for human expression.''
	\citet{gillotte2020copyright} studies copyright infringement in AI-generated artworks and concludes that regarding the training phase ``an engineer may use copyrighted works to train an Al program to
	generate artwork without incurring infringement liability.''
	\citet{hristov2016artificial} considers a separate issue regarding AI and copyright: whether it should be possible to grant copyright to AI-authored works. Current rulings (\cite{copyright-office-ai}) by the U.S. copyright review board state that wholly AI generated works cannot be considered for copyright.

	Memorization of training samples is considered undesirable for many reasons apart from copyright. 
	\citet{LeeINZECC22} show that deduplication can significantly reduce memorization, but not eliminate it (see also bottom row of Table 1 in~\citep{KandpalWR22}).
	\citet{IppolitoTNZJLACC22} state that ``deduplication does not guarantee that a model will not still memorize individual (deduplicated) examples, necessitating defenses that operate at inference-time''. They also show that simply stopping models from outputting training samples verbatim does \emph{not} prevent memorization and can give a ``false sense of security.''
	\citep{LeeLCL22, TirumalaMZA22, CarliniIJLTZ22} show that memorization becomes worse with model size and data reuse. The deterioration with growing model size holds even in the single-epoch (no data reuse) case; see in particular Figures~1 and 8 in \citep{TirumalaMZA22}.

	As discussed in Section~\ref{sec:discussbody}, our work is related to, but also substantially different than, differential privacy~\citep{DworkMNS06}. \citet{Elkin-Koren} study the differences between copyright and privacy anf find that ``if privacy is adopted as standard for copyright infringement, it may undermine copyright law intended purposes''.
	\citet{PonomarevaBV22} train a small (60m parameters) differentially private language model, while \citet{li2021large} fine-tune large models in a differentially private fashion. We discuss additional relevant works on differential privacy in Appendix~\ref{sec:dp}. \citet{carlini2021extracting} show a reconstruction attack of training data from model weights for GPT-2, while very recently \citet{carlini2023extracting} gave training-points reconstruction attacks for diffusion models. 
	We note that while a reconstruction attack has strong privacy implications, it does not prevent copyright protection for the generated outputs.
	
	The work of~\citet{scheffler2022} is closely related to our work. While the goals are different (they analyze prior cases, while we want to build tools to prevent future infringement), the two works are similar on a technical level. Specifically, our Definition~\ref{def:bounded} of 
	$k$-NAF can be interpreted in their framework since $\log(1/\Pr[y])$ for a generative model 
	corresponds to the description length of the randomness used to generate 
	$y$. Plugging this in we get that our parameter $k$ in near access-freeness corresponds to their notion of empirical derivation similarity. Our setup is more directly applicable to generative models due to its probabilistic nature. This is what allows us to give transformations in Section 3 which can ensure $k$-NAF.

	\section{Discussion}~\label{sec:discussion}
	
	This work provided a precise definition for quantifying the extent in which
	a generative-model copies protected
	material.  As discussed, applying our definition in practice requires
	making application-specific choices on the admissible bound $k$, the
	information measure $\Delta$, and ensuring that $\safe(C)$ truly maps
	to a model that did not access $C$.  However, by making these choices
	explicit, we hope this can advance the current state of using, at best,
	heuristic protections against memorizing inputs.  We also hope that
	this work can help to form a basis for discussions between content creators,
	model designers, model users, and legal scholars about the appropriate
	choices.
	
	Our work puts into stark relief the difference between the issues of
	\emph{privacy}, \emph{memorization}, \emph{trademarks}, \emph{patents}, \emph{fair use}, and \emph{copyright}, showing that solution concepts for the latter goal need not address the former goals.
	Indeed, our algorithms use the underlying models as black-boxes, and
	so our resulting model may include a full description of the
	underlying training data it is based on.  In particular, our approach
	makes no attempt to prevent reconstruction of the training-set from
	the model description, as that is unnecessary for investigating
	inference-time copyright infringement. Neither do our algorithms
	attempt to address
	trademark; it may be possible to prompt an LLM go generate material
	that would be considered an infringement of trademark.
	
	Our algorithms are practical, but we believe there is more room for
	optimizations in both training and inference.

	\section*{Acknowledgements and Funding}
	
	We thank the reviewers for their insightful comments.
	\\
	
	\noindent This work has been made possible in part by a gift from the Chan Zuckerberg Initiative Foundation to establish the Kempner Institute for the Study of Natural and Artificial Intelligence.
	Sham Kakade acknowledges funding from the Office of Naval Research under award N00014-22-1-2377 and the National Science Foundation Grant under award \#CCF-2212841.
	Nikhil Vyas  acknowledges funding from NSF grant DMS-2134157 and DOE grant DE-SC0022199.
	Boaz Barak acknowledges funding from a Simons Investigator Fellowship, NSF grant DMS-2134157, DOE grant DE-SC0022199 and DARPA grant W911NF2010021.

	\bibliography{ref}
	
	
\end{document}